\documentclass[10pt,twocolumn,letterpaper]{article}

\usepackage{iccv}
\usepackage{booktabs}
\usepackage{xcolor}
\usepackage{bbding}
\usepackage{enumitem}
\usepackage{multirow}
\usepackage{bbm}
\usepackage{arydshln}
\usepackage{xspace}
\usepackage{cite}
\usepackage{adjustbox}
\usepackage{booktabs}

\usepackage{times}
\usepackage{epsfig}
\usepackage{graphicx}
\usepackage{amsmath}
\usepackage{amssymb}
\usepackage{makecell}
\usepackage{multirow}
\usepackage{float}

\usepackage{graphicx}
\usepackage{amsmath}
\usepackage{amssymb}
\usepackage{booktabs}
\usepackage{xcolor}
\usepackage{bbding}

\usepackage[hang,flushmargin]{footmisc} % align the footprint

\usepackage[pagebackref,breaklinks,colorlinks]{hyperref}

\newcommand{\myparagraph}[1]{{\vspace{.2em} \noindent \bf #1}}

\newcommand*{\affaddr}[1]{#1} % No op here. Customize it for different styles.
\newcommand*{\affmark}[1][*]{\textsuperscript{#1}}

\usepackage[capitalize]{cleveref}
\crefname{section}{Sec.}{Secs.}
\Crefname{section}{Section}{Sections}
\Crefname{table}{Table}{Tables}
\crefname{table}{Tab.}{Tabs.}

\iccvfinalcopy % *** Uncomment this line for the final submission

\def\iccvPaperID{5351} % *** Enter the ICCV Paper ID here

\ificcvfinal\fi

\begin{document}

%%%%%%%%% TITLE
\title{Prompt-ICM: A Unified Framework towards \\ Image Coding for Machines with Task-driven Prompts}

% \and

% \maketitle

\author{Ruoyu Feng\affmark[1*] \quad Jinming Liu\affmark[2*] \quad Xin Jin\affmark[3] \quad Xiaohan Pan\affmark[1] 
Heming Sun\affmark[2]  \quad Zhibo Chen \affmark[1]\textsuperscript{,}\affmark[\dag]\\
\affaddr{\small\affmark[1]University of Science and Technology of China\quad}
\affaddr{\affmark[2]Waseda University\quad} 
\affaddr{\small\affmark[3]Eastern Institute of Advanced Study}
}

\maketitle

\let\thefootnote\relax\footnotetext{*~First two authors contributed equally.\\
 \dag~Corresponding author.}

\ificcvfinal\thispagestyle{empty}\fi

%%%%%%%%% ABSTRACT
\begin{abstract}
Image coding for machines (ICM) aims to compress images to support downstream AI analysis instead of human perception. 
For ICM, developing a unified codec to reduce information redundancy while empowering the compressed features to support various vision tasks is very important, which inevitably faces two core challenges: 
1) How should the compression strategy be adjusted based on the downstream tasks? 2) How to well adapt the compressed features to different downstream tasks? Inspired by recent advances in transferring large-scale pre-trained models to downstream tasks via prompting, in this work, we explore a new ICM framework, termed Prompt-ICM. To address both challenges by carefully learning \textbf{task-driven prompts} to coordinate well the compression process and downstream analysis. Specifically, our method is composed of two core designs: a) \textbf{compression prompts}, which are implemented as importance maps predicted by an information selector, and used to achieve different content-weighted bit allocations during compression according to different downstream tasks; b) \textbf{task-adaptive prompts}, which are instantiated as a few learnable parameters specifically for tuning compressed features for the specific intelligent task. 
Extensive experiments demonstrate that with a single feature codec and a few extra parameters, our proposed framework could efficiently support different kinds of intelligent tasks with much higher coding efficiency.
\end{abstract}

\begin{figure}[htbp]
  \centerline{\includegraphics[width=1.0\linewidth]{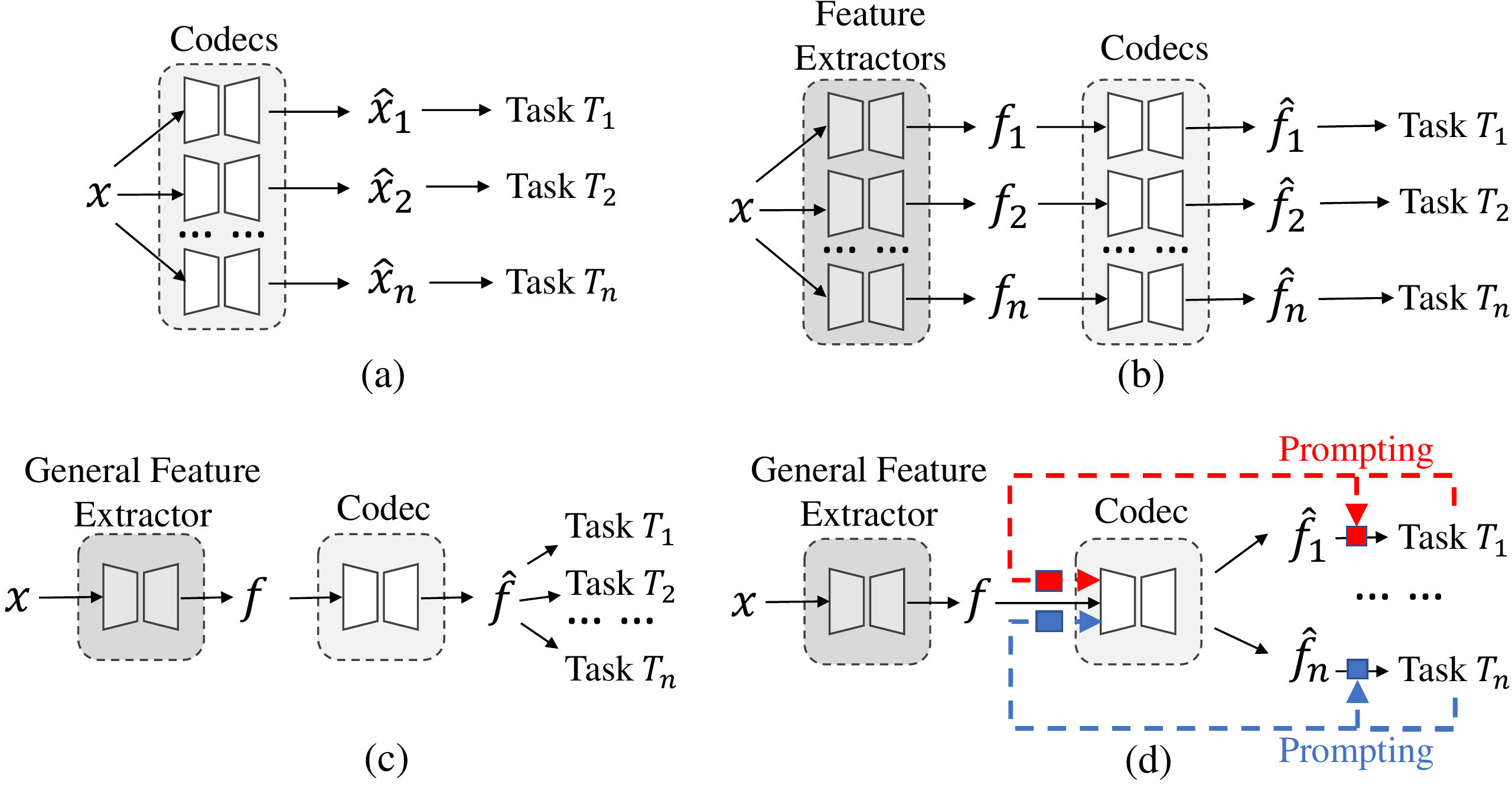}}
    \caption{Different pipelines of image coding for machines (ICM): (a). Using the compressed images to support downstream tasks; (b). One-to-one feature-based ICM pipeline; (c). General features based ICM pipelines which ignores the explicit interaction between the compression task and downstream tasks. (d). Compared with (c). we further consider using task-driven prompts (those colours) to better coordinate the compression process and downstream analysis.} 
\label{fig:motivation}
\end{figure}

%%%%%%%%% BODY TEXT
\section{Introduction}
\label{sec:intro}

In modern society, intelligent multimedia applications have played an irreplaceable role in our daily life, such as smart cities, intelligent surveillance, and the Internet of Things (IoT). 
With the fast development of machine vision technologies, there will be more and more images that need to be compressed and transmitted over the Internet to serve intelligent analysis. One of the key technologies is lossy compression, which aims to save storage resources and transmission bandwidth. 
In the past decades, hand-crafted image and video codecs \cite{wallace1992jpeg,rabbani2002overview,wiegand2003overview,sullivan2012overview,bross2021overview} have significantly improved coding efficiency.

Recently, learned-based codecs \cite{balle2017end,balle2018variational,minnen2018joint, cheng2020learned,johnston2018improved,li2018learning,li2020learning,mentzer2018conditional,mentzer2020high} have shown strong potential, which not only outperform traditional hand-crafted codecs in PSNR, but also can be optimized according to perception-related metrics (e.g., MS-SSIM \cite{wang2004image}, LPIPS\cite{zhang2018unreasonable}) to generate more realistic images.
However, these codecs are mainly designed to satisfy human perception. 
When facing AI task analysis, existing image coding methods (even the learned-based ones) are still questionable. 
% Since the information requirements of intelligent tasks and human vision are quite different, and there exist various even unknown tasks, 
% employing the existing codecs to compress images for downstream tasks would invariably produce sub-optimal results.
Due to the fundamental differences between the information needs of intelligent tasks and human vision, and the existence of various, perhaps even unknown tasks, utilizing existing codecs to compress images for downstream tasks is likely to yield suboptimal outcomes.

Therefore, a new task of compressing images for machine vision, called \textbf{image coding for machines (ICM)} \cite{duan2020video,le2021image}, has emerged to build a joint efficient and analytical framework. 
Such a framework is capable of obtaining and compressing general representations to effectively support intelligent analytics for massive and diverse applications.

The existing ICM methods can be divided into three branches: 
Figure \ref{fig:motivation} (a) shows the first branch that uses task-specific codecs to compress images \cite{le2021image,le2021learned}, and then perform intelligent analysis based on reconstructed images. 
These codecs are typically optimized for their respective task losses along with rate losses, in an end-to-end manner. 
As shown in Figure \ref{fig:motivation} (b), methods in the second branch \cite{duan2015overview,duan2018compact,chen2019lossy,chen2019toward,singh2020end,ma2018joint,babenko2014neural} firstly extract specific features for individual compression, and finally use the reconstructed features to complete the intelligent task analysis. 
Note that, such two branches have a large limitation: the different intelligent tasks need to use their corresponding codecs for compression, respectively, \ie, lack of generalization.
And the lack of generalization might cause a significant extra cost of computation and storage for different downstream tasks.
To overcome this defect, the third branch has been explored (as shown in Figure \ref{fig:motivation} (c) \cite{feng2022image}). 
% It is composed of a general feature extractor and its corresponding feature codec. 
% Inputs of all downstream tasks are the reconstructed general features. 
% Nevertheless, this kind of method is still not that efficient because it does not take into account the task-specific characteristics during compression and downstream analysis, or said, the relationship between optimized compression scheme and each downstream supporting task has not been well exploited.
This method comprises a generic feature extractor and its associated feature codec, which reconstructs the general features for all subsequent tasks. However, it suffers from suboptimal efficiency due to its disregard for task-specific characteristics during compression and downstream analysis. In other words, the potential benefits of an optimized compression scheme tailored to individual downstream tasks have not been fully leveraged.

In this paper, we tend to explore a new framework for image coding for machines (ICM) that circumvents the aforementioned issues.
Inspired by the recent successes of parameter efficient tuning for transferring large-scale pre-trained models to downstream tasks~\cite{jia2022visual,nie2022pro,chen2022conv,zhang2021tip,zhou2022learning,zhou2022conditional}, we design a new ICM framework, termed as Prompt-ICM, from a new aspect of carefully learning task-driven prompts to coordinate well the compression process and downstream analysis, as shown in Figure \ref{fig:motivation} (d).
This framework consists of two core designs.
The first design is compression prompts, which refer to importance maps that represent the positional importance distribution conditioned on the extracted features and the corresponding intelligent task.
More specifically, the compression prompts are predicted by a lightweight information selector (IS) module and utilized in conjunction with a spatially variable-rate feature compression model to achieve content-weighted bit allocation during the compression process, tailored to the specific task requirements.
The second component of our framework is task-adaptive prompts, which incorporate a few additional learnable parameters to analyze the compressed features for the specific downstream task. 
Together with compression prompts, enabling our Prompt-ICM framework to utilize a unified codec that efficiently supports diverse intelligent tasks with superior compression performance.

The main contributions of this paper are summarised as follows:

\begin{itemize}[label=$\bullet$]

    \item To the best of our knowledge, we are the first to investigate and formulate the coordination of the interaction between compression and downstream analytics in a unified framework. Our proposed Prompt-ICM can support different kinds of intelligent tasks based on only a single codec.
    \item We propose the compression prompts for content-weighted compression according to the demands of downstream tasks. As a by-product contribution, we design an effective sub-component, a lightweight information selector (IS) module, to predict importance maps as compression prompts.
    \item Furthermore, we propose task-adaptive prompt tuning to transfer compressed features for downstream tasks, achieving significant performance improvement with a few parameters, which is more practical for ICM applications.

\end{itemize}

\section{Related Work}
\label{sec:related_work}
\subsection{Image Compression}
Image compression aims to represent original pixel samples using a compact and high-fidelity format. 
 Traditional hand-crafted image codecs typically involve intra prediction, discrete cosine transformation or wavelet transformation, quantization, and entropy coding \cite{wallace1992jpeg,rabbani2002overview,wiegand2003overview,sullivan2012overview,bross2021overview}.
Learned-based codecs \cite{balle2017end,balle2018variational,minnen2018joint, cheng2020learned,johnston2018improved,li2018learning,li2020learning,mentzer2018conditional,mentzer2020high} make use of neural networks to learn to minimize distortion between pairs of source images and reconstructed
images, while maximizing the likelihood of the quantized
latent representation for low bitrate in an end-to-end manner. 
% Besides, an important advantage of learned-based compression models is the versatility gained by joint optimization of perceptual metrics, such as MS-SSIM\cite{wang2004image}, LPIPS\cite{zhang2018unreasonable}, and adversarial loss\cite{mentzer2020high}. 
% Compression models optimized with these metrics generate more realistic images, though with sub-optimal signal fidelity. 
% However, since the rate-distortion trade-off is controlled by a Lagrange multiplier $\lambda$, most existing methods are limited in that a fixed value of $\lambda$ corresponds to a single point in the rate-distortion curve. 
Furthermore, the utilization of learned-based compression models offers a significant advantage in terms of versatility through the joint optimization of perceptual metrics such as MS-SSIM\cite{wang2004image}, LPIPS\cite{zhang2018unreasonable}, and adversarial loss\cite{mentzer2020high}. Despite a potential decrease in signal fidelity, compression models optimized with these metrics can produce more realistic images. 
However, since the rate-distortion trade-off is controlled by a Lagrange multiplier $\lambda$, most existing methods are limited in that a fixed value of $\lambda$ corresponds to a single point in the rate-distortion curve. 
Recent works\cite{choi2019variable,cui2020g,johnston2018improved,toderici2017full,yang2020variable} propose different approaches to support variable rates using a single model. 
Song \etal \cite{song2021variable} propose to perform spatial bit allocation according to a quality map that is the same size as the original image. 

\subsection{Image Coding for Machines}
Image coding for machines (ICM) targets at compressing and transmitting source images to support downstream intelligent tasks, such as image classification\cite{jia2022visual,he2016deep,liu2021swin,han2021transformer,dosovitskiy2020image}, object detection\cite{ren2015faster,redmon2016you, redmon2017yolo9000,lin2017feature,li2022exploring}, instance segmentation\cite{he2017mask,liu2018path,bolya2019yolact}, and semantic segmentation\cite{long2015fully,badrinarayanan2017segnet,chen2017deeplab,chen2018encoder,xie2021segformer,zheng2021rethinking}. A natural way is joint optimization \cite{akbari2019dsslic,hu2020towards,li2021task,le2021image,wang2021towards} of image compression models and the downstream intelligent tasks. 
Another branch of intuitive methods compresses the features\cite{duan2015overview,duan2018compact,chen2019lossy,chen2019toward,singh2020end,ma2018joint,babenko2014neural} of corresponding tasks instead of images for both coding efficiency and computing offloading. 
Recently, Feng \etal\cite{feng2022image} propose to learn features that are both general and compact based on joint optimization of self-supervised learning and entropy constraint.
And all intelligent tasks are performed based on the extracted features. 
Nevertheless, this method doesn't consider the coordination between the compression process and downstream transferring, lacking targeted adjustments for different tasks. 
Differently, this paper aims to design a unified framework that contains the advantages of the above methods and avoids the corresponding disadvantages. 
More specifically, we explore the coordination between general feature compression and downstream task transferring and propose a unified framework that can adapt to different kinds of machine vision tasks based on a single compression model with a few learnable parameters.

\subsection{Parameter Efficient Tuning for Large-scale Pre-trained Models}
Parameter efficient tuning (PET) is first introduced in NLP \cite{houlsby2019parameter,pfeiffer2020adapterhub,li2021prefix,lester2021power,liu2021pre,he2021towards} since it's inefficient to fully fine-tune all parameters of large-scale pre-trained models \cite{brown2020language,radford2018improving,radford2019language,radford2021learning,jia2021scaling} on each downstream intelligent task.
In computer vision, parameter efficient tuning is first introduced to large-scale pre-trained visiaon-language  models\cite{radford2021learning,jia2021scaling} via prompt-based tuning\cite{zhou2022learning,zhou2022conditional}, which introduces additional learnable prompts attached to the input during the training stage, while keeping the pre-trained models fixed. 
Zhang \etal\cite{zhang2021tip} and Gao \etal \cite{gao2021clip} design lightweight adapters
to predict the adapted feature residuals to modulate representation space. 
Jia \etal \cite{jia2022visual} adapt visual prompts for supervised pre-trained vision transformers. Bahng \etal \cite{bahng2022exploring} explore visual prompts in input pixel space for adapting pre-trained models. 
Nie \etal \cite{nie2022pro} inserts several lightweight prompt blocks into backbones to adjust feature representation.
This paper considers a more practical scenario of multiple downstream intelligent tasks supported for ICM. 
In combination with PET methods, our framework can support different downstream tasks more efficiently.

\begin{figure*}[htbp]
  \centerline{\includegraphics[width=1.0\linewidth]{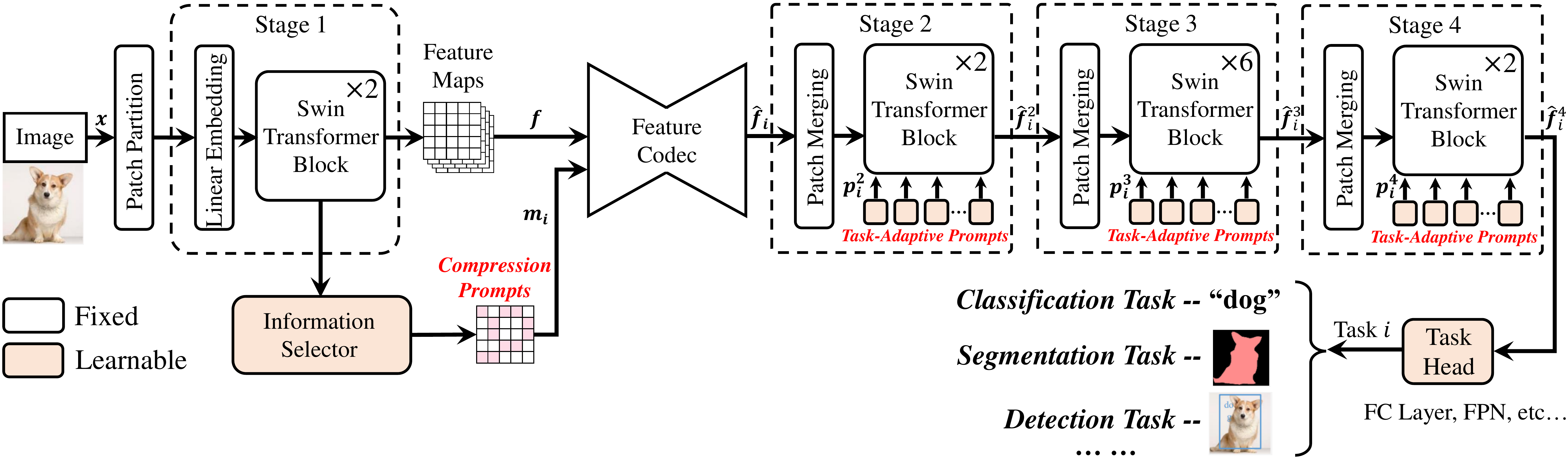}}
  
    \caption{
    The framework of our proposed Prompt-ICM (taking Swin-T as an example). Downstream transferring is performed via task-driven prompt tuning. 
    By tuning the lightweight information selector and task-adaptive prompts, Prompt-ICM could efficiently support various downstream tasks, e.g., classification, segmentation, and detection.
    }
\label{fig:main_framework}
\end{figure*}

\section{Approach}
\label{sec:method}
% \subsection{\color{red}{Expressions in Mathematical Form}}
\subsection{Formulation of General ICM}
In this section, we mathematically define the problem of general ICM from a new perspective, which is formulated by Equation (\ref{equ:1})-(\ref{equ:2}).
To begin with, the input image $\boldsymbol{x}$ is firstly analyzed by the pre-trained feature extractor $\boldsymbol{FE}_i$ with parameters $\theta_{\mathrm{\boldsymbol{FE}_\textit{i}}}$ to extract the feature $\boldsymbol{f}_i$ for task $i$:
\begin{equation}
\boldsymbol{f}_i=\boldsymbol{FE}_i\left(\boldsymbol{x} ; \,\theta_{\mathrm{\boldsymbol{FE}_\textit{i}}}\right).
\label{equ:1}
\end{equation}
Note that when $\boldsymbol{FE}_i$ is None, $\boldsymbol{f}_i$ refers to the raw image $\boldsymbol{x}$ for subsequent operations, as Figure \ref{fig:motivation} (a) shows.
\par
After that, a lossy codec $\boldsymbol{C}_i$ with parameters $\theta_{\boldsymbol{C}_i}$ is used to compress the features $\boldsymbol{f}_i$ for task $i$:

\begin{equation}
\hat{\boldsymbol{f}}_i=\boldsymbol{C}_i\left(\boldsymbol{f}_i ; \,\theta_{\boldsymbol{C}_i}\right).
\label{equ:codec}
\end{equation}

Then the reconstructed feature $\hat{\boldsymbol{f}}_i$ is sent to the remaining networks $\boldsymbol{T}_i$ with parameters $\theta_{\boldsymbol{T}_i}$ to acquire prediction results $o_i$:

\begin{equation}
o_i=\boldsymbol{T}_i\left(\hat{\boldsymbol{f}}_i ;\, \theta_{\boldsymbol{T}_i}\right).
\label{equ:predict}
\end{equation}

Generally, the optimization function of the ICM framework for downstream transferring can be described as:

\begin{equation}
\begin{aligned}
    &argmin_{\Phi = \left\{\theta_{\boldsymbol{FE}_i}, \theta_{\boldsymbol{C}_i}, \theta_{\boldsymbol{T}_i}\right\}} \alpha \mathcal{L}_i+R. \\
% &subject\ to: K_{({\Phi})}<K_t,
\end{aligned}
\label{equ:2}
\end{equation}
where the Lagrange multiplier $\alpha$ controls the trade-off between bitrate $R$ and loss $\mathcal{L}_i$ for task $i$. 
% At the same time, different from previous ICM methods, we consider a more practical scenario of limiting the number of parameters $K_{({\Phi})}$ learnable to the downstream transferring task $i$ to be smaller than the required number of parameters $K_t$.

\subsection{Overview}
In contrast to general ICM, we propose a new ICM framework called Prompt-ICM. Firstly, similar to Figure \ref{fig:motivation} (c), we use a single \textbf{general feature extractor} with fixed parameters $\varphi_{\boldsymbol{FE}}$\footnote[1]{In this paper, we use $\theta$ to represent learnable parameters, while using $\varphi$ to represent fixed parameters for downstream tasks.} to extract the general feature for all downstream tasks instead of extracting different features for different tasks. The Equation (\ref{equ:1}) in our framework can be revised as follows:
\begin{equation}
\boldsymbol{f}=\boldsymbol{FE}\left(\boldsymbol{x} ; \,\varphi_{\mathrm{\boldsymbol{FE}}}\right).
\label{equ:6}
\end{equation}
This will significantly reduce the extra cost of computation and storage for different downstream tasks.
\par
However, the framework corresponding to Figure \ref{fig:motivation} (c) does not take into account the task-specific characteristics during compression, which may result in inefficient coding for specific tasks. 
To mitigate the aforementioned issue, we propose a lightweight information selector module ($\boldsymbol{IS}$) with tunable parameters $\theta_{\boldsymbol{IS}}$. 
% To address this problem, we design a lightweight information selector (denoted as $\boldsymbol{IS}$) module with parameters $\theta_{\boldsymbol{IS}}$ to predict importance maps as compression prompts $\boldsymbol{m}$. 
The module generates importance maps as \textbf{compression prompts}, which are used to guide the spatial bit allocation of the codec. 
% And then, we adopt \textbf{compression prompts} to guide the spatially bit allocation of the codec. 
The general feature extractor and customized compression prompts enable us to employ a single controllable feature codec $\boldsymbol{C}$ with parameters $\varphi_{\boldsymbol{C}}$ for all various downstream tasks instead of designing distinct codecs for different tasks. 
Therefore, Equation (\ref{equ:codec}) in our framework can be revised as:
\begin{equation}
\label{equ:IS}
\boldsymbol{m}_i=\boldsymbol{IS}_i\left(\boldsymbol{f} ; \,\theta_{\boldsymbol{IS}_i}\right).
\end{equation}
\begin{equation}
\hat{\boldsymbol{f}}_i=\boldsymbol{C}\left(\boldsymbol{f}, \boldsymbol{m}_i ; \,\varphi_{\boldsymbol{C}}\right).
\end{equation}
\par 
The next step is to send the reconstructed feature $\hat{\boldsymbol{f}}_i$ to $\boldsymbol{T}_i$ to acquire prediction. 
Nevertheless, fine-tuning $\boldsymbol{T}_i$ for each downstream task is parameter-consuming. 
In this paper, we utilize the \textbf{task-adaptive prompts} $\boldsymbol{p}_i$ with a few learnable parameters $\theta_{\boldsymbol{p}_i}$ to conduct efficient transferring. 
Then the Equation (\ref{equ:predict}) in our framework can be revised as follows:
\begin{equation}
o_i=\boldsymbol{T}_i\left(\hat{\boldsymbol{f}}_i, \boldsymbol{p}_i  ; \,\varphi_{\boldsymbol{T}^{'}_i}, \theta_{{\boldsymbol{p}_i}}, \theta_{{\boldsymbol{h}_i}}\right).
\label{equ:decoder}
\end{equation}
where the parameters of $\boldsymbol{T}_i$ are divided to two parts: $\theta_{{\boldsymbol{h}_i}}$ denotes the parameters of the task head, while remaining fixed parameters are represented as $\varphi_{\boldsymbol{T}^{'}_i}$. 
\par
Notably, when transferring to downstream tasks, we only need to fine-tune the information selector, the task head, and task-adaptive prompts. The optimization function can be revised as follows:
\begin{equation}
\begin{aligned}
    &argmin_{\Phi^{'} = \left\{\theta_{\boldsymbol{IS}_i}, \theta_{{\boldsymbol{h}_i}}, \theta_{\boldsymbol{p}_i}\right\}} \alpha \mathcal{L}_i+R, \\
% &subject\ to: K_{({\Phi})}<K_t,
\end{aligned}
\label{equ:7}
\end{equation}
where the number of trainable parameters $\Phi^{'}$ in Equation (\ref{equ:7}) is far fewer than $\Phi$ in Equation (\ref{equ:2}).
% Before downstream transferring by Prompt-ICM, we first train a controllable feature codec as preparation.
% The feature codec is trained to compress general features extracted from the large-scale pre-trained model. 
% More importantly, the feature codec can spatially conduct bit allocation according to the compression prompts. 
% When it comes to the downstream transferring by task-driven prompt tuning, our Prompt-ICM includes both the compression prompts in the encoder side and task-adaptive prompts in the decoder side. 
% During downstream transferring, compression prompts are predicted by a lightweight information selector (IS) and input together with features to the well-trained compression model. 
% Compression prompts indicate the importance of individual regions in the feature and serve as a guide for the compression model to perform task-aware content-weight compression.
% And task-adaptive prompts are introduced to precisely tune features during the forward propagation process, efficiently tuning features towards downstream task.
% The whole network is optimized in an end-to-end manner, with only parameters of the information selector and prompts learnable, resulting in an unified and efficient framework for heterogeneous tasks.  

\begin{figure}[htbp]
  \centerline{\includegraphics[width=1.0\linewidth]{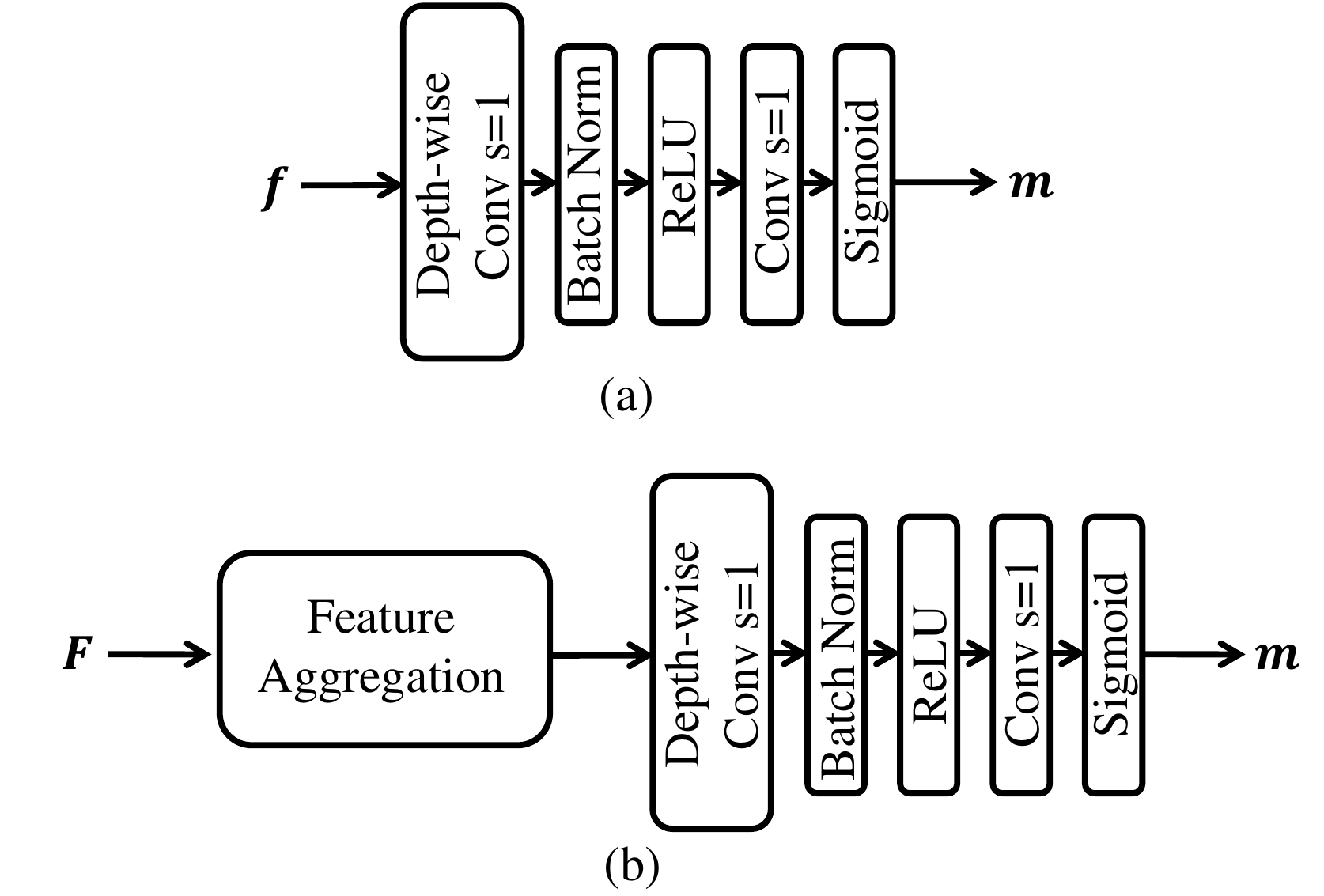}}
    \caption{Two variants of information selector modules that differ in the input features for generating compression prompts.
    }
\label{fig:information_selector}
\end{figure}

\subsection{General Feature Extraction}
\label{sec:approach_feature_extraction}
With a large-scale pre-trained vision model, the whole network is first divided into multiple sub-layers as stages. 
We follow the regular concepts of stage partitioning\cite{he2016deep,liu2021swin}. 
Considering that the features would be consumed by a variety of kinds of intelligent tasks, \eg, image classification, object detection, and semantic segmentation, the features extracted at stage 1 (with a 4x down-sampling factor) are taken as the general features to promise completeness of information and integrity of the content's spatial layout. 
% Formally, consider the stage 1 $s_{1}$ from an $n$-stage pre-trained vision model $\boldsymbol{S}=\{s_{i}\}$.  An input image $\boldsymbol{x}$ is fed into $s_{1}$, obtaining an $4\times$ down-sampling feature, formulated as:
Formally, consider $s_{1}$ to be the first stage of a pre-trained vision model $\boldsymbol{S}=\{s_{j}\}_{j=1}^{n}$ with $n$-stages.
% , where $ j\in\{1, 2, \ldots, n\}$. 
Given an input image $\boldsymbol{x}$, we feed it into $s_{1}$ to obtain the feature map $\boldsymbol{f}$ with 4$\times$ down-sampling, i.e., $\boldsymbol{FE} = s_1$.
% denoted as:

% \begin{equation}
%     \small
%     \boldsymbol{f}=s_{1}(\boldsymbol{x}).
% \end{equation}

In this paper, we take the Swin Transformer\cite{liu2021swin} as the base model for its strong representation capability and functionalities resulting from the hierarchical design. 

\subsection{Compression Prompts}
\label{sec:approach_feature_compression}
During each inference, we use compression prompts $\boldsymbol{m}$ generated by a lightweight information selector ($\boldsymbol{IS}$) to guide content-weighted feature compression corresponding to the current task.

\myparagraph{Generation of Compression Prompts}. 
As shown in Figure \ref{fig:information_selector} (a), $\boldsymbol{IS}$ module is used to extract importance maps as compression prompts. 
Besides, we can perform additional forward propagation at the encoding side to obtain multi-scale features $\boldsymbol{F}=\{\boldsymbol{f}^k\}_{k=1}^{n}$ which contain richer and hierarchical information. Note that $\boldsymbol{f}^1$ in $\boldsymbol{F}$ corresponds to $\boldsymbol{f}$ in Section \ref{sec:approach_feature_extraction}.
Then, the Equation (\ref{equ:IS}) can be revised as $\boldsymbol{m}_i=\boldsymbol{IS}_i\left(\boldsymbol{F} ; \,\theta_{\boldsymbol{IS}_i}\right)$.
% $, $\boldsymbol{f} = \boldsymbol{F}$
And the $\boldsymbol{IS}$ module can aggregate these features from multiple semantic levels to better generate compression prompts, as shown in Figure \ref{fig:information_selector} (b). 
% formulated as: 
% \begin{equation}
%     \boldsymbol{m}_i = \boldsymbol{IS}_i(\boldsymbol{F}|\theta_{\boldsymbol{IS}_i}).
% \end{equation}

\myparagraph{Content-weighted Feature Compression}. 
To make use of the compression prompts, we design a controllable feature codec by adjusting previous learned lossy compression methods to enable compression prompts $\boldsymbol{m}_i$ to guide content-weighted feature compression for task $i$. 
% It is important to note that the training of the codec is independent of the rest of the network in the whole framework to ensure the irrelevance of any specific task. 
Notably, to ensure the irrelevance of any specific task, the training of the codec is independent of the rest of the network in the whole framework.
\par
In previously learned lossy image compression, the goal is to simultaneously minimize the bitrate and the distortion. Such an objective can be formulated as minimizing:
\vspace{-1mm}
\begin{equation}
    R+{\lambda}D.
    \label{equ:rdo}
\end{equation}
\vspace{-1mm}
where $R$ represents the bitrate, $D$ denotes the distortion between the original features and the reconstructed features, and $\lambda$ is the Lagrange multiplier that controls the rate-distortion trade-off. 
In our framework, we go beyond previous learned-based codecs in that one model controlled by a fixed value of $\lambda$ corresponds to a single point in the rate-distortion curve, and build a feature codec that can conduct bit allocation according to manually set compression prompts $\boldsymbol{m}$. 
% Different from simple variable-rate compression, our feature codec implicitly conducts bit allocation during compression according to the compression prompts $\boldsymbol{m}$.
% Compression prompts are tensors of the same spatial size as the input features but with only one channel.
% Values of compression prompts denote the importance of corresponding positions, relative to bits allocated to those pixels, in the range [0,1].
Thus the Equation (\ref{equ:rdo}) is newly written as:
\vspace{-1mm}
\begin{equation}
    R+{\boldsymbol{\Lambda}}\cdot{\boldsymbol{D}},
\end{equation}
\vspace{-1mm}
where ${\boldsymbol{\Lambda}}=\left\{{\lambda}_{h,w}\right\}_{h=1,w=1}^{H,W}$ denotes the importance of each position, and $\boldsymbol{\Lambda} = \boldsymbol{m}$.
 $\boldsymbol{D}=\left\{{D}_{h,w}\right\}_{h=1,w=1}^{H,W}$ represents the distortion in each position of the feature $\boldsymbol{f}$ and the reconstructed feature $\hat{\boldsymbol{f}}$. 

More specifically, we design the compression framework derived from the Mean \& Scale (M\&S) Hyperprior model \cite{minnen2018joint} and spatial variable-rate image compression \cite{song2021variable}. 
Since round-based quantization is non-differential, the additive uniform noise\cite{balle2017end} is added to the latent variables for rate estimation during training.

The overall R-D (rate-distortion) loss function for the training of the codec is formulated as:
\vspace{-1mm}
\begin{equation}
\label{eq:feature_compression_loss_fn}
\begin{aligned}
% \begin{split}
    \mathcal{L}_f =& \mathcal{R}(\hat{\boldsymbol{y}}) + \mathcal{R}({\hat{\boldsymbol{z}}}) + \boldsymbol{\Lambda}\cdot{\boldsymbol{D}}(\boldsymbol{f},\hat{\boldsymbol{f}})\\
    =&\mathop{\mathbb{E}}[-\log_2(p_{\hat{\boldsymbol{y}}|\hat{\boldsymbol{z}}}({\hat{\boldsymbol{y}}|\hat{\boldsymbol{z}}}))] + \mathop{\mathbb{E}}[-\log_2(p_{\hat{\boldsymbol{z}}|{\boldsymbol{\psi}}}({\hat{\boldsymbol{z}}|{\boldsymbol{\psi}}}))] \\
    &+ \sum_{h=1}^{H}\sum_{w=1}^{W}{\lambda}_{h,w}\frac{(f_{h,w}-\hat{f}_{h,w})^{2}}{HW},
% \end{split}
\end{aligned}
\end{equation}
where $p_{\hat{\boldsymbol{y}}|\hat{\boldsymbol{z}}}(\hat{\boldsymbol{y}}|\hat{\boldsymbol{z}})$ denotes the probability distribution of the latent variable $\hat{\boldsymbol{y}}$ which is a compact representation of $\boldsymbol{f}$ and generated by the encoder of the codec. $\hat{\boldsymbol{y}}$ and side information $\hat{\boldsymbol{z}}$ which provides hyper-prior information for $\hat{\boldsymbol{y}}$ are encoded as bitstreams for transmission and storage. The decoded bitstreams are used to generate the reconstructed feature $\hat{\boldsymbol{f}}$. $p_{\hat{\boldsymbol{z}}|{\boldsymbol{\psi}}}(\hat{\boldsymbol{z}}|{\boldsymbol{\psi}})$ denotes the probability distribution of side information $\hat{\boldsymbol{z}}$, $\boldsymbol{\psi}$ denotes the factorized density model to encode $\hat{\boldsymbol{z}}$, $H$ and $W$ denote the height and width of the feature, and ${\lambda}_{h,w}$ denotes the Lagrange multiplier of the corresponding position. 
The detailed network architecture, compression process, and formulations are reported in the supplementary.
\par 
After the training stage of the feature codec, its parameters are fixed. When transferring to downstream task $i$, 
the task-oriented adjustment is achieved by fine-tuning the lightweight $\boldsymbol{IS}_i$ for generating $\boldsymbol{m}_i$, resulting in a unified codec for various intelligent tasks.
% $\boldsymbol{m}_i$ is generated by the $\boldsymbol{IS}_i$.
\begin{figure*}[htbp]
  \centering
  \medskip
    \centerline{\includegraphics[width=1.0\linewidth]{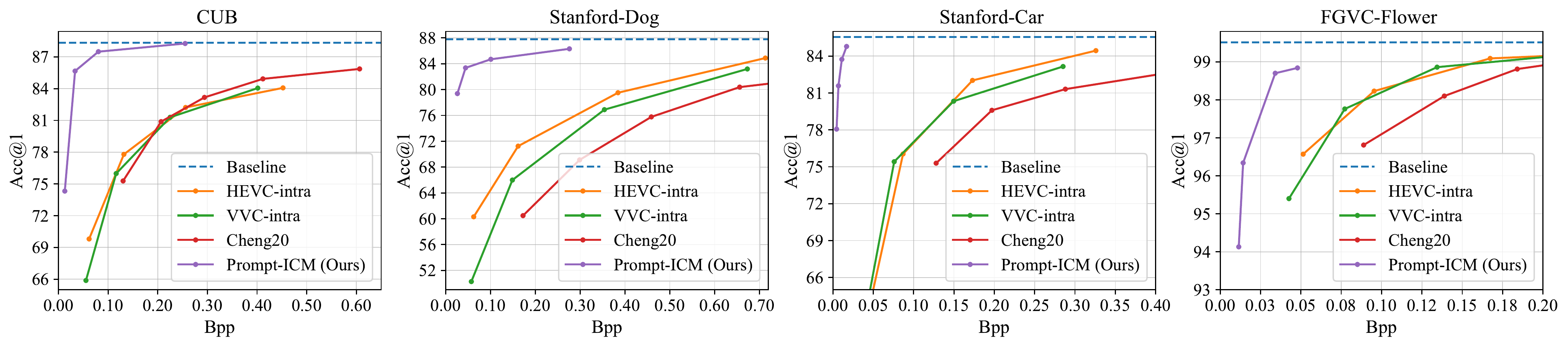}}
    \vspace{-1.5mm}
  \caption{Classification results on different datasets at various bitrates. Two traditional codecs HEVC-intra \cite{sullivan2012overview}, VVC-intra \cite{bross2021overview}, and one learned-based codec \emph{Cheng20}~\cite{cheng2020learned} are compared with our method. 
%   We evaluate these methods with TAPT (top) and full tuning (bottom).
  }
\label{fig:cls_swin_b_rate_acc}
\end{figure*}
\vspace{-5mm}
\begin{figure*}[htbp]
  \centerline{\includegraphics[width=1\linewidth]{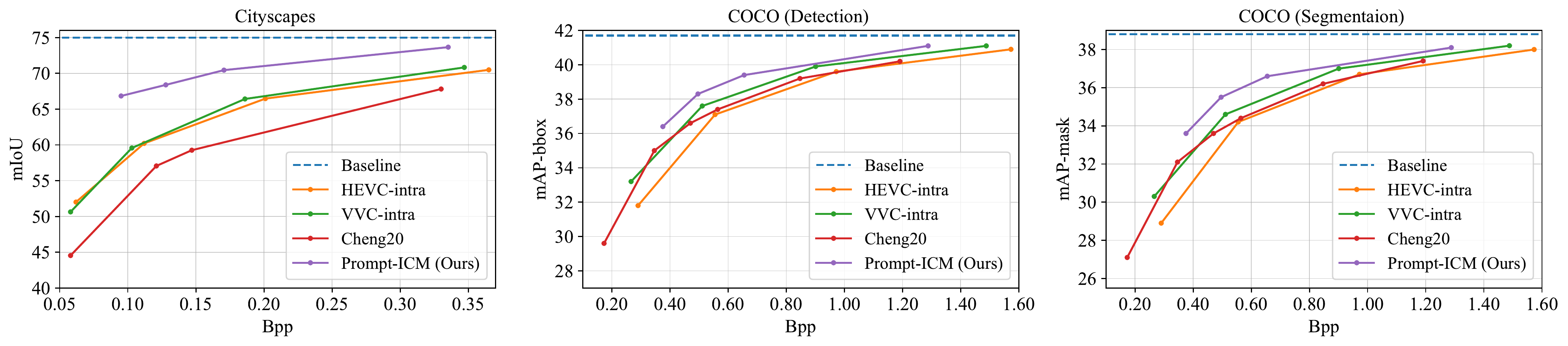}}
    \vspace{-1.5mm}
    \caption{Results of semantic segmentation on Cityscapes (the first one) and object detection and instance segmentation on COCO 2017 (second and third).
     }
     \vspace{-5mm}
\label{fig:coco_swin_t_rate_map}

\end{figure*}

\vspace{4mm}
\subsection{Task-adaptive Prompts}
Task-adaptive prompts are instantiated as a few learnable parameters specifically for tuning compressed features for image analysis. They are injected into the pre-trained models on the decoder side, and fine-tuned to fit the specific downstream tasks. 
It should be noted that the parameters of task-adaptive prompts are much smaller than those of the original task model.
\par
As Figure \ref{fig:main_framework} shows, after obtaining the reconstructed feature $\hat{\boldsymbol{f}}_i$ of task $i$, 
task-adaptive prompts $\boldsymbol{p}_i=\{\boldsymbol{p}_{i}^k\}_{k=2}^{n}$ are introduced to adjust features during the forward propagation in the rest \textit{n}-1 stages, corresponding to Equation (\ref{equ:decoder}).

The overall loss function for transferring to downstream task $i$ is given by:
\vspace{-1mm}
\begin{equation}
\vspace{-1mm}
    \mathcal{L} = \mathcal{R}(\hat{\boldsymbol{y}}) + \mathcal{R}({\hat{\boldsymbol{z}}}) + \alpha
    \mathcal{L}_{i}(o_i, {gt}_i),
\end{equation}
where the $\mathcal{R}(\hat{\boldsymbol{y}})$ and $\mathcal{R}({\hat{\boldsymbol{z}}})$ denote rate of the latent variables $\hat{\boldsymbol{y}}$ and side information $\hat{\boldsymbol{z}}$, 
% $\mathcal{L}_{i}(\cdot,\cdot)$ denotes the loss of the current task, $o_i$ denotes the output for task $i$, ${gt}_i$ denotes the ground truth,
$\mathcal{L}_{i}(\cdot,\cdot)$, $o_i$, and ${gt}_i$ denote the loss function, output, and ground truth of the current task, respectively, and $\alpha$ is the Lagrange multiplier to achieve the trade-off between the task loss and bitrates. Note that in the downstream transferring, only parameters of the information selector, task-adaptive prompts, and the task head are learnable, while the feature extractor, feature compression model, and pre-trained stages are all fixed, thus achieving efficient downstream task transferring.

Thanks to compression prompts for content-weighted information selection and task-adaptive prompt tuning, Prompt-ICM achieves both coding efficiency and parameter efficiency for the downstream transfer to heterogeneous tasks with only a single feature codec, resulting in a simple yet unified framework for image coding for machines. 

\section{Experiments}
\label{sec:experiments}
\subsection{Datasets}
For training of the feature codec, we use ImageNet\cite{deng2009imagenet} as the training database. 
As for the verification of downstream task transferring, we experiment on four image classification datasets and two dense prediction datasets.
The four image classification datasets are CUB-200-2011\cite{wah2011caltech}, Stanford Dogs\cite{khosla2011novel}, Stanford Cars\cite{gebru2017fine}, and Oxford Flowers\cite{nilsback2008automated}, respectively.
The two datasets for dense prediction are COCO 2017\cite{lin2014microsoft} and Cityscapes\cite{cordts2016cityscapes}. 
COCO 2017 is a dataset for dense prediction tasks of object detection and instance segmentation that contains 118K training images, 5K validation images, and 20K test-dev images.
Cityscapes is a fundamental and challenging dataset, specifically for semantic segmentation.
It has 5,000 high-quality images with pixel-level annotations in total, with 2975 for training, 500 for validation, and 1525 for testing, respectively.

\begin{figure}[t]
  \vspace{-2mm}
  \centerline{\includegraphics[width=0.95\linewidth]{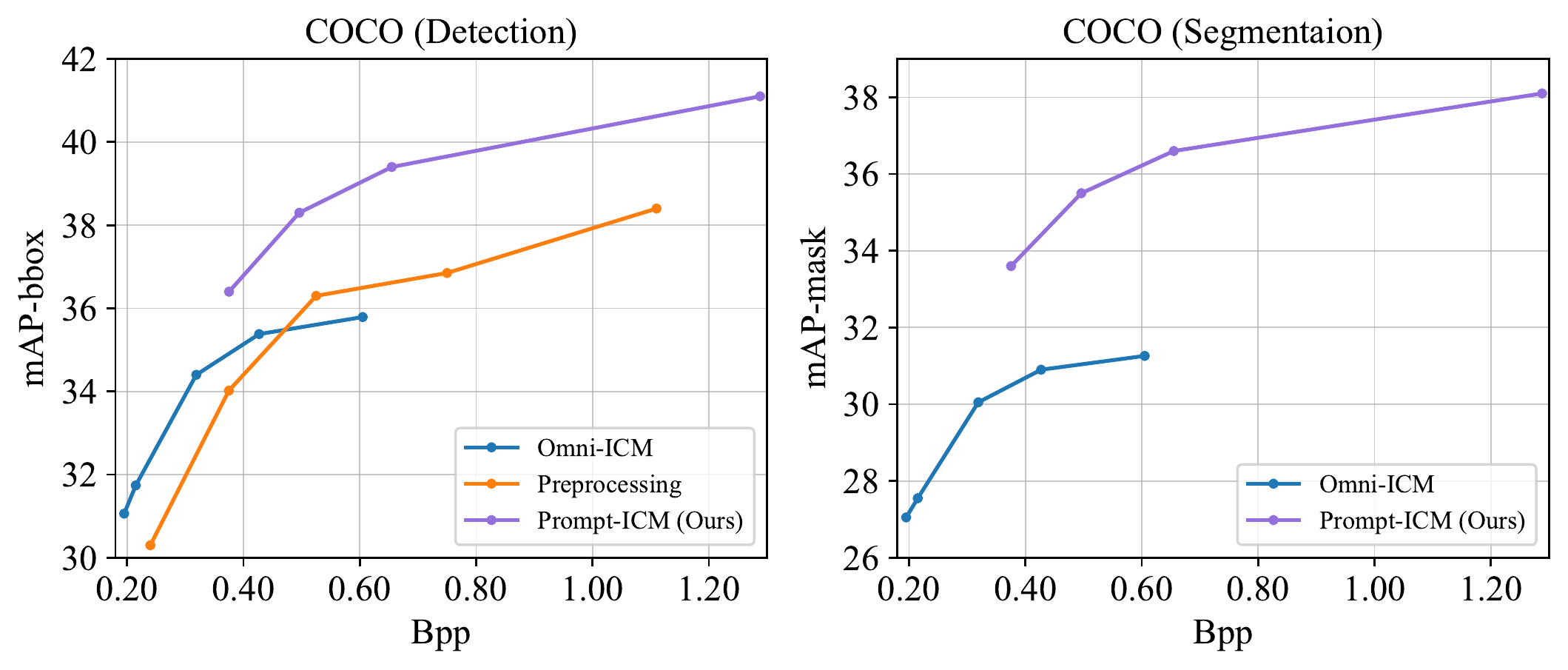}}
    \vspace{-2mm}

    \caption{Comparisons with SOTA ICM methods in terms of object detection and instance segmentation on COCO 2017.
     }
     \vspace{-2mm}
\label{fig:comparison_sota}

\end{figure}

\subsection{Implementation Details}
\myparagraph{Large-scale Pre-trained Backbones.} 
% We conduct experiments with the Swin Transformer\cite{liu2021swin} pre-trained on ImageNet-21K\cite{deng2009imagenet}. 
% Experiments conducted on image classification are mainly based on Swin-Base and experiments conducted on dense prediction are based on Swin-Tiny.
% Besides, we also conduct experiments with Vision Transformer (ViT) \cite{dosovitskiy2020image}, and results of experiments performed under ViT are included in the supplementary.
We performed experiments using the Swin Transformer\cite{liu2021swin} model pre-trained on the ImageNet-21K dataset\cite{deng2009imagenet}. For image classification experiments, we used the Swin-Base model, while for dense prediction experiments, we used the Swin-Tiny model. Additionally, we conducted experiments using the Vision Transformer (ViT) \cite{dosovitskiy2020image}, and the results are included in the supplementary.

\myparagraph{Controllable Feature Compression.}
We train the controllable feature codec for 2M iterations with a batch size of 8. Adam\cite{kingma2014adam} optimizer is employed, and the learning rate is 3e-5 and decreases to 3e-6 after 1.8M iterations. 
The manually set compression prompts $\boldsymbol{m}$, i.e.,  ${\boldsymbol{\Lambda}}$ in Equation (\ref{eq:feature_compression_loss_fn}) is uniformly sampled from [0.5, 32], resulting in a bpp range of [0.02, 1.0] on the Kodak dataset. 
During the training stage of the codec, to ensure the variety of possible compression prompts, we randomly generate each instance in a mini-batch by using one of the four different ways (1) a uniform map (2) a gradation map between two randomly selected values (3) a kernel density estimation map of a Gaussian mixture with random mean, variance, and a number of mixtures (4) a map consisting of various blocks in a grid manner.

\myparagraph{Downstream Transferring via Task-Driven Prompt Tuning.} 
For compression prompts, we take multi-scale feature aggregation as input to the information selector by default. 
For task-adaptive prompts, visual prompt tuning (VPT) \cite{jia2022visual} is instantiated for image classification, and Pro-Tuning \cite{nie2022pro} is instantiated for dense prediction, \ie object detection, instance segmentation, and semantic segmentation. 
We follow the default settings in their original papers. 

\subsection{Effectiveness and Superiority}
\myparagraph{Evaluation Protocol}. 
% We evaluate the rate-distortion performance on different kinds of intelligent tasks. 
% The distortion part of rate-distortion is represented by metrics of corresponding intelligent tasks.
% And the rate part is measured by the bits per pixel (bpp), which is formulated as $\frac{b}{h \times w}$, 
% where $h, w$ are the height and width of the source image, 
% and $b$ denotes the total bits cost of the coded feature bit-stream.
We evaluate the rate-distortion performance across various intelligent tasks. The distortion component is represented by metrics specific to each task. The rate component is determined by bits per pixel (bpp), which is computed as $\frac{b}{h \times w}$, where $h$ and $w$ denote the height and width of the source image, respectively, and $b$ refers to the total bits utilized by the coded feature bitstream. 

\myparagraph{Comparison Approaches}. 
We mainly compare our method with the most advanced codecs, including traditional codecs (HEVC\cite{sullivan2012overview}, VVC\cite{bross2021overview}) and a learned-based codec \emph{Cheng20}~\cite{cheng2020learned}. 
We take the results obtained by feeding uncompressed raw images into the task model as the baseline, or said, performance upper bound.
% For all the evaluations of the compared approaches, we feed reconstructed images into the well-trained task model (that is trained with uncompressed raw images) to obtain corresponding results. 
For all subsequent evaluations of the compared approaches, reconstructed images are fed into the task model, which has been previously trained with uncompressed raw images, to obtain the corresponding results.

\begin{table}[t]
\small
\centering
\setlength{\tabcolsep}{1mm}

\caption{The comparison of learnable parameters between the method of Prompt-ICM and the method of full tuning on different tasks.}
\begin{tabular}{c|c|c|c}
% \hline
% \toprule[1pt]
\Xhline{1.0pt}
\textbf{\begin{tabular}[c]{@{}c@{}}Trainable\\ Parmeters (M)\end{tabular}} & \multicolumn{1}{c|}{\textbf{\begin{tabular}[c]{@{}c@{}}Cla.\\ (Swin-B)\end{tabular}}} & \multicolumn{1}{c|}{\textbf{\begin{tabular}[c]{@{}c@{}}Det. \& Ins.\\ (Swin-T)\end{tabular}}} & \multicolumn{1}{c}{\textbf{\begin{tabular}[c]{@{}c@{}}Sem.\\ (Swin-T)\end{tabular}}} \\ 
% \hline \hline
% \hline
\Xhline{0.5pt}
% \midrule[1pt]
% \textbf{w/o CP}
%                         & 0.24                                                                                           & 24.68                                                                                      & 36.83                                                                                        \\ 
\textbf{Full Tuning}                                                   & 87.61                                                                                           & 47.49                                                                                      & 59.64                                                                                        \\ 
\textbf{Prompt-ICM}                                                           & 0.87                                                                                           & 25.31                                                                                      & 37.46     \\ 
% \hline                     
% \bottomrule[1pt]
\Xhline{1.0pt}
\end{tabular}
\label{tab1:parameters_comparison}
\end{table}

\begin{figure}[t]
  \centerline{\includegraphics[width=0.7\linewidth]{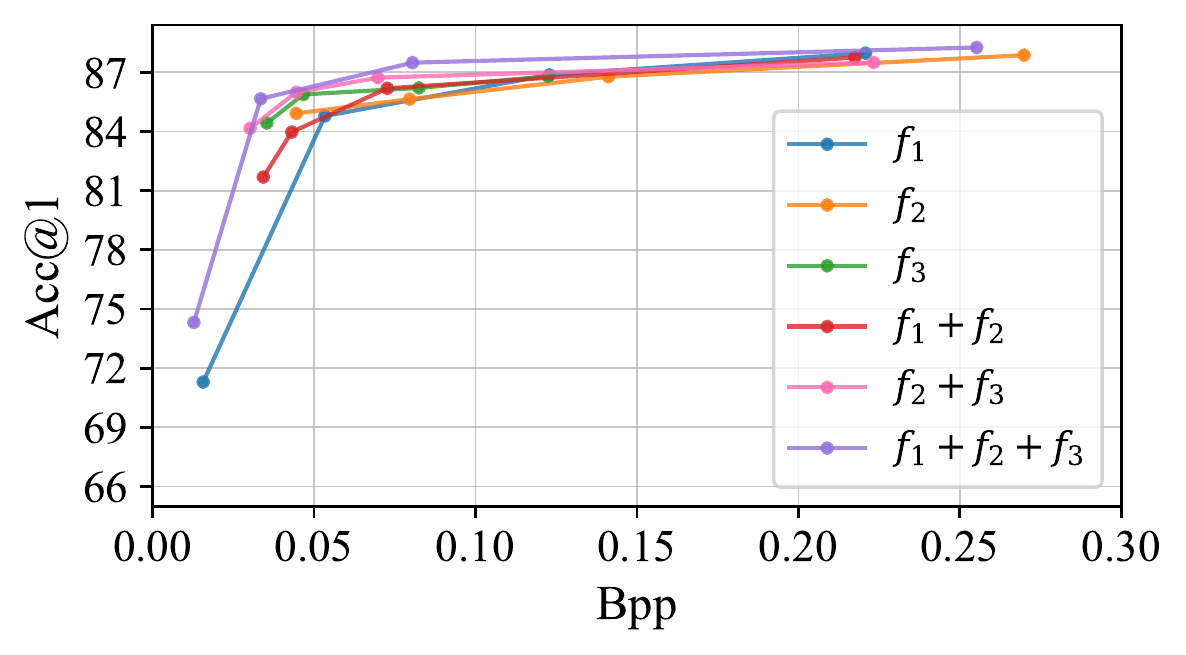}}
    \vspace{-2mm}

    \caption{
    Ablation studies on generation of compression prompts using features of different stages. 
     }
     \vspace{-2mm}
\label{fig:feature_generator_ablation_study}
\end{figure}

% \vspace{-3mm}

% \vspace{-6mm}

\begin{figure}[t]
  \centerline{\includegraphics[width=1.0\linewidth]{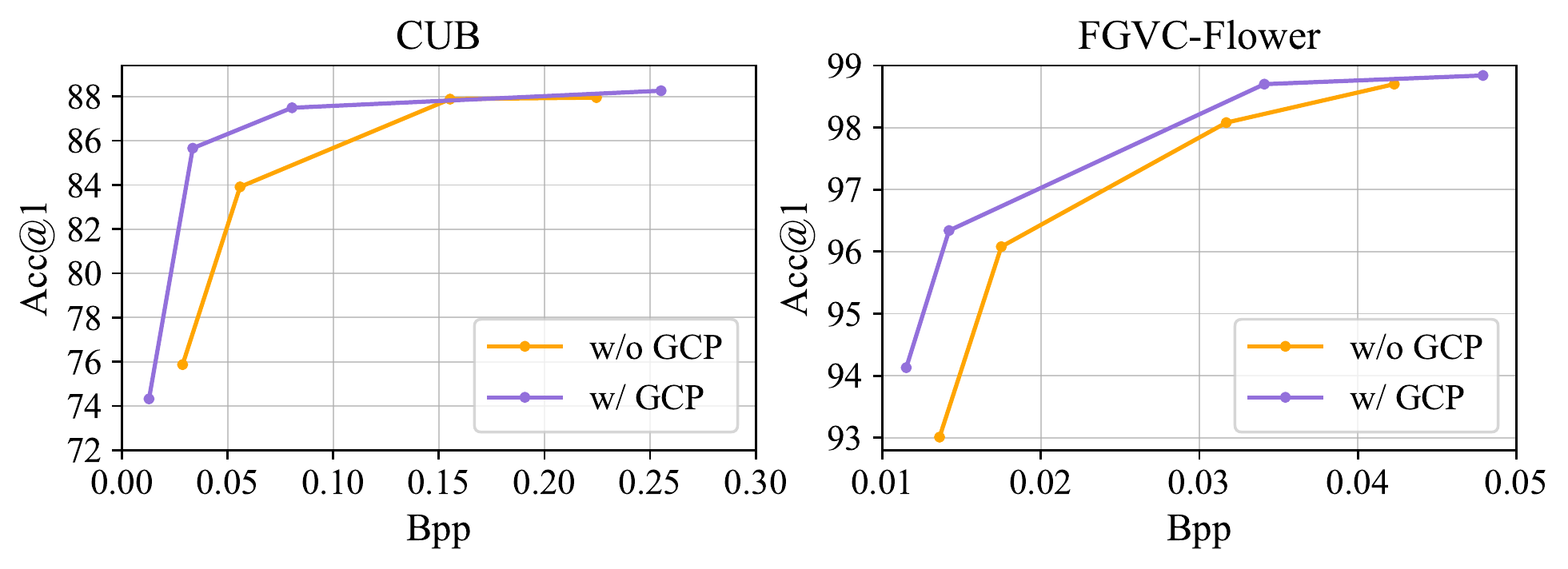}}
    \vspace{-2mm}

    \caption{Ablation studies on generated compression prompts (GCP) based on different datasets. ``w/ GCP" represents using generated compression prompts to guide content-weighted coding, while ``w/o GCP" refers to the coding with manual compression prompts and cannot allocate bits adaptively. 
    }
    
\label{fig:information_selection_ablation}
\vspace{-4mm}
\end{figure}

% \vspace{-1mm}

% \vspace{-1mm}

\myparagraph{Image Classification}. 
For training, we follow the default optimization settings in \cite{jia2022visual}.
When evaluating classification tasks, we resize and crop each input image to $224 \times 224$ before inputting them into the model. 
As shown in Figure \ref{fig:cls_swin_b_rate_acc}, the R-D performance of our method exceeds compared methods by a significant margin. 
Surprisingly, our method can perform well at extremely low bitrates (0.03$\sim$0.1 bpp).
Meanwhile, as shown in Table \ref{tab1:parameters_comparison}, Prompt-ICM only requires 0.87M learnable parameters to be updated during transferring, while the full-tuning scheme requires 87.61M learnable parameters.
Last but not least, all results of Prompt-ICM are achieved by a unified feature codec, which is critical to practical application scenarios. 

\myparagraph{Dense Prediction}. 
For semantic segmentation on Cityscapes, we utilize UperNet\cite{xiao2018unified} implemented in mmseg\cite{mmseg2020} as the base framework.
AdamW\cite{loshchilov2017decoupled} optimizer with the learning rate of 6e-5 and a weight decay of 0.01 is employed. 
We use a batch size of 16 for 80K training iterations with the crop size of 512${\times}$512. 
For object detection and instance segmentation tasks on COCO 2017, Mask R-CNN\cite{he2017mask} with FPN\cite{zhao2017pyramid} is utilized as the detector implemented in mmdet\cite{mmdetection}. 
We follow the common protocol that the image scale is in [800, 1333] pixels during both the training and inference stages by default.
AdamW\cite{loshchilov2017decoupled} optimizer (initial learning rate of 1e-4, weight decay of 0.05, and batch size of 16) is used with 12 epochs. 
As shown in Figure \ref{fig:coco_swin_t_rate_map}, Prompt-ICM outperforms all other methods on the three dense prediction tasks.
Thanks to the conditional compression prompts that help with the content-weighted compression process, Prompt-ICM can allocate more bits to those task-related regions for the dense prediction tasks, which is further confirmed in Section \ref{sec:visual}.
Meanwhile, as shown in Table \ref{tab1:parameters_comparison}, the required numbers of learnable parameters for object detection and instance segmentation, and semantic segmentation are 25.31M and 37.46M, while those of full tuning are 47.49M and 59.64M. It shows that our task-adaptive prompts enable the proposed Prompt-ICM framework to achieve significant parameter savings when transferring to dense prediction tasks.
\begin{figure}[t]
  \vspace{-3mm}
  \centerline{\includegraphics[width=1.0\linewidth]{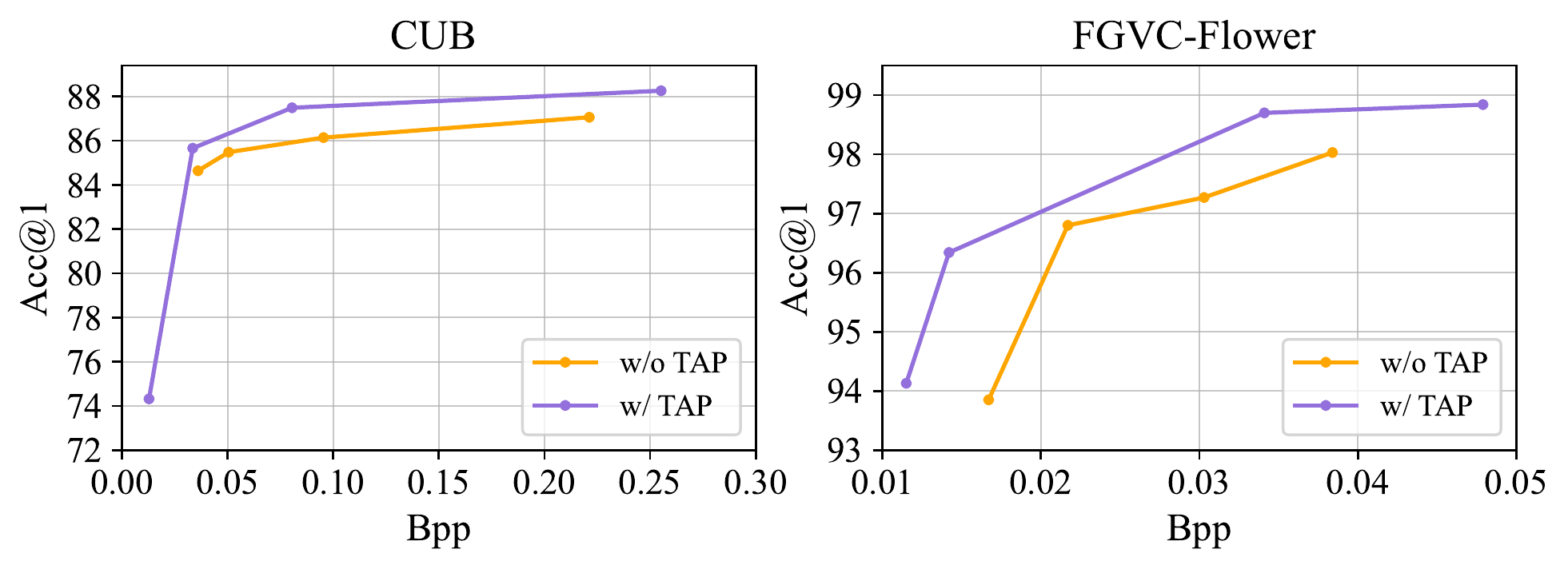}}
    \vspace{-1mm}

    \caption{
    Ablation studies on task-adaptive prompts (TAP). ``w/o TAP'' corresponds to fully tuning all parameters of pre-trained backbones and the task head on the decoder side.
    }
    \vspace{-2mm}
\label{fig:cls_swin_b_full_rate_acc}
\end{figure}

\begin{figure*}[htbp]
  \vspace{-3.5mm}
  \centerline{\includegraphics[width=1.0\linewidth]{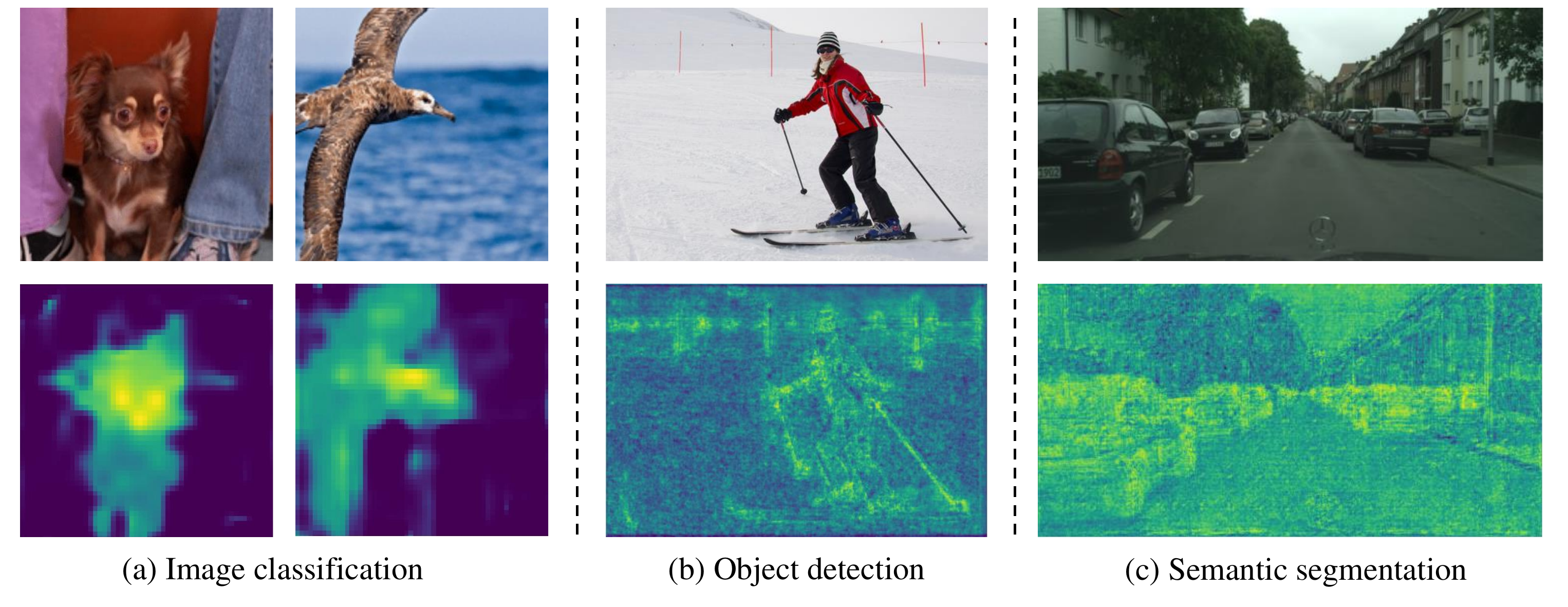}}
    \caption{
    Visualisation results of compression prompts on different tasks, including image classification, object detection, and semantic segmentation. Positions with higher brightness in the compression prompts mean that they are more important. 
     }
\label{fig:classification_visual}

\end{figure*}

\myparagraph{Comparison with SOTA ICM Methods.} We compare our proposed method with Omni-ICM\cite{feng2022image} and preprocessing scheme~\cite{lu2022preprocessing}, which also only use a single codec for completing ICM tasks.
As shown in Figure \ref{fig:comparison_sota}, our method achieves the best performance and significantly outperforms others with much fewer learnable parameters.
Additional comparisons on other datasets are reported in the supplementary due to limited space.

\subsection{Ablation Study}

\subsubsection{Study on Compression Prompts}
\label{sec:exp_cp}
\myparagraph{Generation of Compression Prompts}. 
% We test the effect of using different combination of features to generate compression prompts for classification
% and the results are shown in Figure \ref{fig:feature_generator_ablation_study}. 
% First, integrating all features from stage 1 to stage 3 obtains the best performances, which indicates that the information from multiple semantic levels are all helpful for importance localization.
% Besides, it can also be inferred that no matter what kind of combination of features is used, the performance will not change much. 
% The most suitable combination can be selected according to the computing power and requirements.
% The present study examines the impact of various combinations of features utilized for generating compression prompts in classification. 
We study the effect of various combinations of features for generating compression prompts on the classification of the CUB-200-2011 dataset.
As illustrated in Figure \ref{fig:feature_generator_ablation_study}, it can be inferred that the inclusion of all features from stage 1 to stage 3 results in the most superior performance. 
This observation implies that information derived from multiple semantic levels is advantageous for localizing importance. 
Moreover, the results indicate that the performance remains relatively stable regardless of the feature combination employed. 
Thus, the choice of a particular combination should be based on the available computing power and specific requirements to balance the trade-off between performance and complexity. 

% we can consistently achieve a comparable performance without the aggregation when the bitrate is high. The most suitable combination can be selected according to the computing power and requirements.

\myparagraph{Content-weighted Feature Compression}.
To verify that our generated compression prompts adaptively conduct content-weighted feature compression for a specific task, we compare our method to the scheme without using the generated compression prompts. 
By manually setting the compression prompts to a value between 0 and 1 instead of the compression prompts generated by the information selector, we can implement a codec without information selection. Figure \ref{fig:information_selection_ablation} shows that our generated compression prompts lead to a significant improvement in rate-distortion performance on different datasets.   

\vspace{-4mm}
\subsubsection{Study on Task-adaptive Prompts}
Figure \ref{fig:cls_swin_b_full_rate_acc} further presents the ablation study about task-adaptive prompts. 
Combined with the study on learnable parameters shown in Table \ref{tab1:parameters_comparison}, we can infer that task-adaptive prompt tuning achieves even better performances than the scheme of full tuning (w/o TAP), while tuning the task-adaptive prompts only needs a few parameters (0.87M for task-adaptive prompt tuning vs. 87.61M for full tuning) to be updated during downstream transferring. 
These findings further provide evidence that our proposed Prompt-ICM approach possesses excellent properties of both coding efficiency and parameter efficiency.

\subsection{Vision Analysis and Insights}
\label{sec:visual}
We visualize the compression prompts for different tasks as shown in Figure \ref{fig:classification_visual}. 
It can be inferred that compression prompts are mainly concentrated on objects and edges that are closely related to the current task.
During the compression process, compression prompts instruct the codec to allocate more bits to those important regions and fewer bits to less important ones. 
More specifically, for the classification task, heads of dogs and birds are more critical to classification results, while human legs, regions of road, trees, and the sea are unimportant to task inference.
This phenomenon is reasonable and intuitive.
For dense prediction tasks, including semantic segmentation and object detection tasks, the importance is broader and more concentrated on the boundaries of objects, which are essential to precise localization and identification. 
By jointly observing the visualization results of different tasks, it can be inferred that the information selector pays different degrees of attention to tasks of different granularities, pays more attention to discriminative patterns for image-level tasks, and pays more attention to local details for dense prediction tasks.

%-------------------------------------------------------------------------

\section{Conclusion}
\label{sec:conclusion}
We present Prompt-ICM, a unified framework that makes use of large-scale pre-trained models to support a variety of downstream intelligent tasks.
By introducing compression prompts to guide feature compression and task-adaptive prompts for compressed feature tuning, Prompt-ICM can well transfer to different intelligent tasks based on only one feature codec.
Our experiments demonstrate the significant superiority of our framework in a wide range of vision-intelligent tasks.

\clearpage

\clearpage

%%%%%%%%% REFERENCES

{\small
\bibliographystyle{ieee_fullname}
\bibliography{reference}
}

\newpage
\appendix

\section{Controllable Feature Compression}

\begin{figure*}[htbp]
  \centerline{\includegraphics[width=1.0\linewidth]{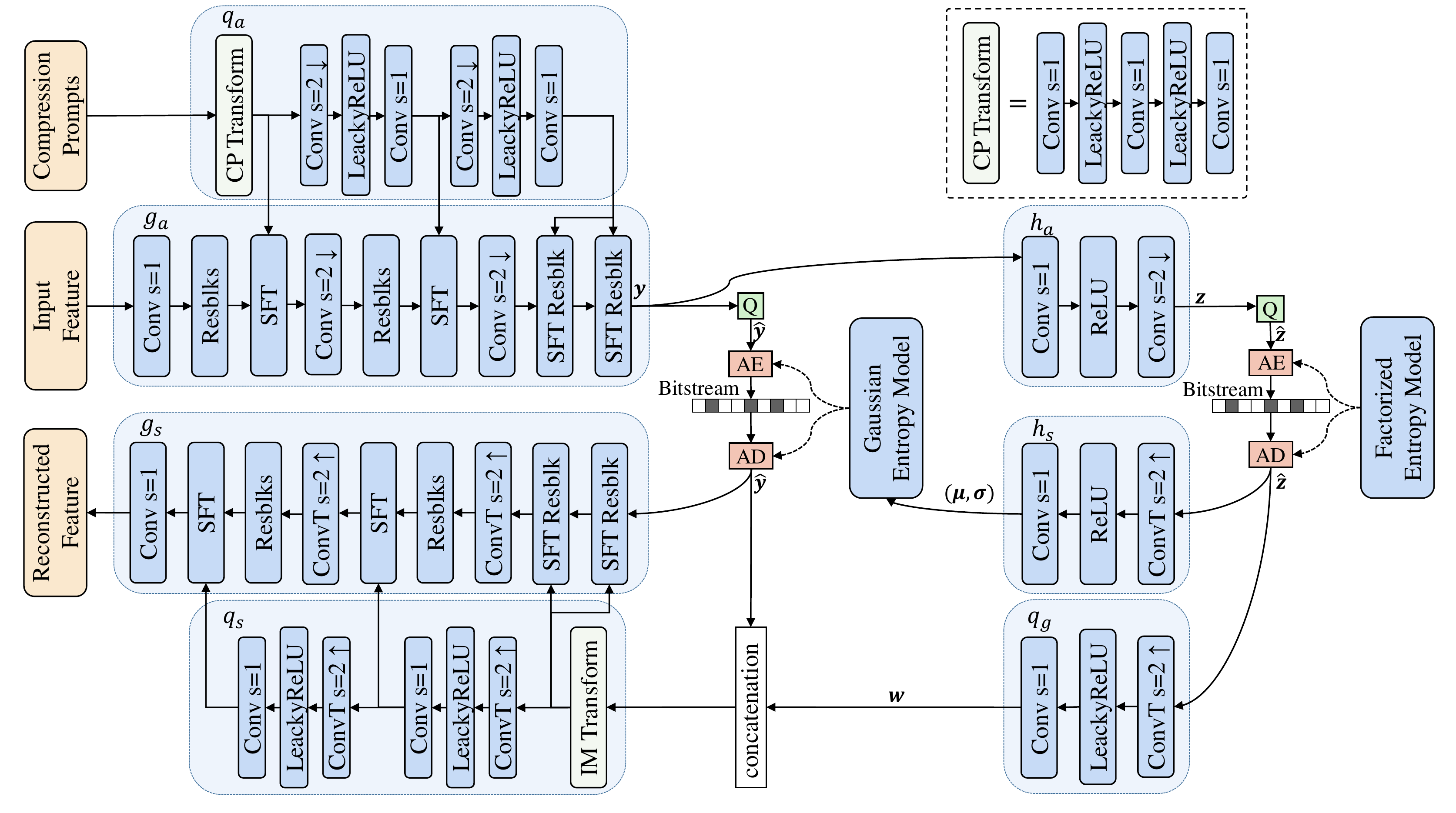}}
    \caption{The architecture of our spatially variable-rate feature compression network.
     }
\label{fig:feature_codec_swin}
\end{figure*}

\myparagraph{Spatially Variable-Rate Feature Compression}. 
As shown in Figure \ref{fig:feature_codec_swin}, the feature $\boldsymbol{f}$ is input to the encoder $g_a$ and the corresponding compression prompt $\boldsymbol{m}$ is input to the condition network $q_a$, obtaining the latent variable $\boldsymbol{y}$. Then quantization is performed. The process can be written by:
\begin{equation}
\begin{split}
    \boldsymbol{y}=g_{a}(\boldsymbol{f},{\Psi}_{a}) ,&\quad \text{where}\quad{\Psi}_{a}=q_{a}(\boldsymbol{m}),\\
    &\hat{\boldsymbol{y}} = Q(\boldsymbol{y}),
\end{split}
\end{equation}
where the $\boldsymbol{m}$ denotes compression prompts, and $Q$ denotes the quantizer.

Since the hard rounding quantization operation of $Q$ is non-differential, an additive noise alters quantization \cite{balle2017end} during training. 
As for the inference stage, after real round-based quantization, entropy coding techniques (e.g., Huffman coding and arithmetic coding\cite{rissanen1981universal}) can losslessly compress the quantized discrete latent variable $\hat{\boldsymbol{y}}$ if the probability distribution $p_{\hat{\boldsymbol{y}}|\hat{\boldsymbol{z}}}(\hat{\boldsymbol{y}}|\hat{\boldsymbol{z}})$ is given. And we use $\hat{\boldsymbol{y}}$ to denote both $\hat{\boldsymbol{y}}$ of the hard quantized latent variable and $\tilde{\boldsymbol{y}}$ of the noised latent variable for simplicity.

Next, the latent variable $\boldsymbol{y}$ is input into the hyper-encoder $h_{a}$ and the quantizer $Q$, obtaining the side-information. This process is formulated as:
\begin{equation}
\begin{split}
    \boldsymbol{z} &= h_{a}(\boldsymbol{y}), \\
    \hat{\boldsymbol{z}} &= Q(\boldsymbol{z}).
\end{split}
\end{equation}

Additive noise is also performed for $\boldsymbol{z}$ during training as an alternative of real round-based quantization for differentiability. And entropy estimation of $\hat{\boldsymbol{z}}$ is performed by a learned factorized entropy prior $\boldsymbol{\psi}$, formulated as:

\begin{equation}
\begin{aligned}
&p_{\hat{\boldsymbol{z}}|{\boldsymbol{\psi}}}(\hat{\boldsymbol{z}}|{\boldsymbol{\psi}}) = \prod_{i} (p_{z_{i}|{\boldsymbol{\psi}}}({\boldsymbol{\psi}})\ast \mathcal{U}(-\frac{1}{2}, \frac{1}{2}))({\hat{z}}_i),
\end{aligned}
\end{equation}
where $z_i$ denotes the $i$-th element of $\boldsymbol{z}$, and $i$ specifies to the position of each signal.

Then the side-information $\hat{\boldsymbol{z}}$ contains both the hyper prior for estimating probability distributions of latent variable $\boldsymbol{y}$ and the conditioned information. 
It is then fed into the hyper-decoder $h_s$ and the condition generator $q_g$, which can be written as:

\begin{equation}
\begin{split}
    p_{\hat{\boldsymbol{y}}|\hat{\boldsymbol{z}}}(\hat{\boldsymbol{y}}|\hat{\boldsymbol{z}}) &\gets h_{s}(\hat{\boldsymbol{z}}),\\
    \boldsymbol{w} &= q_{g}(\hat{\boldsymbol{z}}),
\end{split}
\end{equation}
where $p_{\hat{\boldsymbol{y}}|\hat{\boldsymbol{z}}}(\hat{\boldsymbol{y}}|\hat{\boldsymbol{z}})$ denotes the estimated distribution conditioned on $\hat{\boldsymbol{z}}$ and $\boldsymbol{w}$ represents the spatial conditioned information for feature reconstruction. More specifically, the conditional probability distribution $p_{\hat{\boldsymbol{y}}|\hat{\boldsymbol{z}}}(\hat{\boldsymbol{y}}|\hat{\boldsymbol{z}})$ after decoding $\hat{\boldsymbol{z}}$ is modeled by a mean and scale  Gaussian distribution, which is:

\begin{equation}
    p_{\hat{\boldsymbol{y}}|\hat{\boldsymbol{z}}}(\hat{\boldsymbol{y}}|\hat{\boldsymbol{z}})\sim{\mathcal{N}(\boldsymbol{{\mu}},{\boldsymbol{\sigma}}^2)}.
\end{equation}

% For feature reconstruction, the discrete latent variable $\hat{\boldsymbol{y}}$ is input to decoder $g_{s}$, $\hat{\boldsymbol{y}}$ and the spatial condition information $\boldsymbol{w}$
For feature reconstruction, the decoder $g_{s}$ and condition network $q_{s}$ operates on the latent variable $\hat{\boldsymbol{y}}$ and spatial conditional information $\boldsymbol{w}$, which can be formulated as:

\begin{equation}
    \hat{\boldsymbol{f}}=g_{s}(\hat{\boldsymbol{y}},{\Psi}_{s}) ,\quad \text{where}\quad{\Psi}_{s}=q_{s}(\boldsymbol{w}).
\end{equation}

\myparagraph{Implementation Details}. 
As illustrated in Figure \ref{fig:feature_codec_swin}, we design the compression framework derived from the Mean \& Scale (M\&S) Hyperprior model \cite{minnen2018joint} and spatial variable-rate image compression \cite{song2021variable}. 
% The detailed architecture of our feature compression model is illustrated in Figure \ref{fig:feature_codec_swin}. 
Residual blocks are used to increase the receptive field and representation capability\cite{cheng2020learned}. Besides, the spatial feature transform (SFT) blocks and spatial feature transform residual block (SFT Resblk) shown in Figure \ref{fig:feature_codec_swin} are derived from \cite{song2021variable} to modulate features during non-linear transform process.

\section{Feature Aggregation Information Selector}
\begin{figure}[t]
  \centerline{\includegraphics[width=1.0\linewidth]{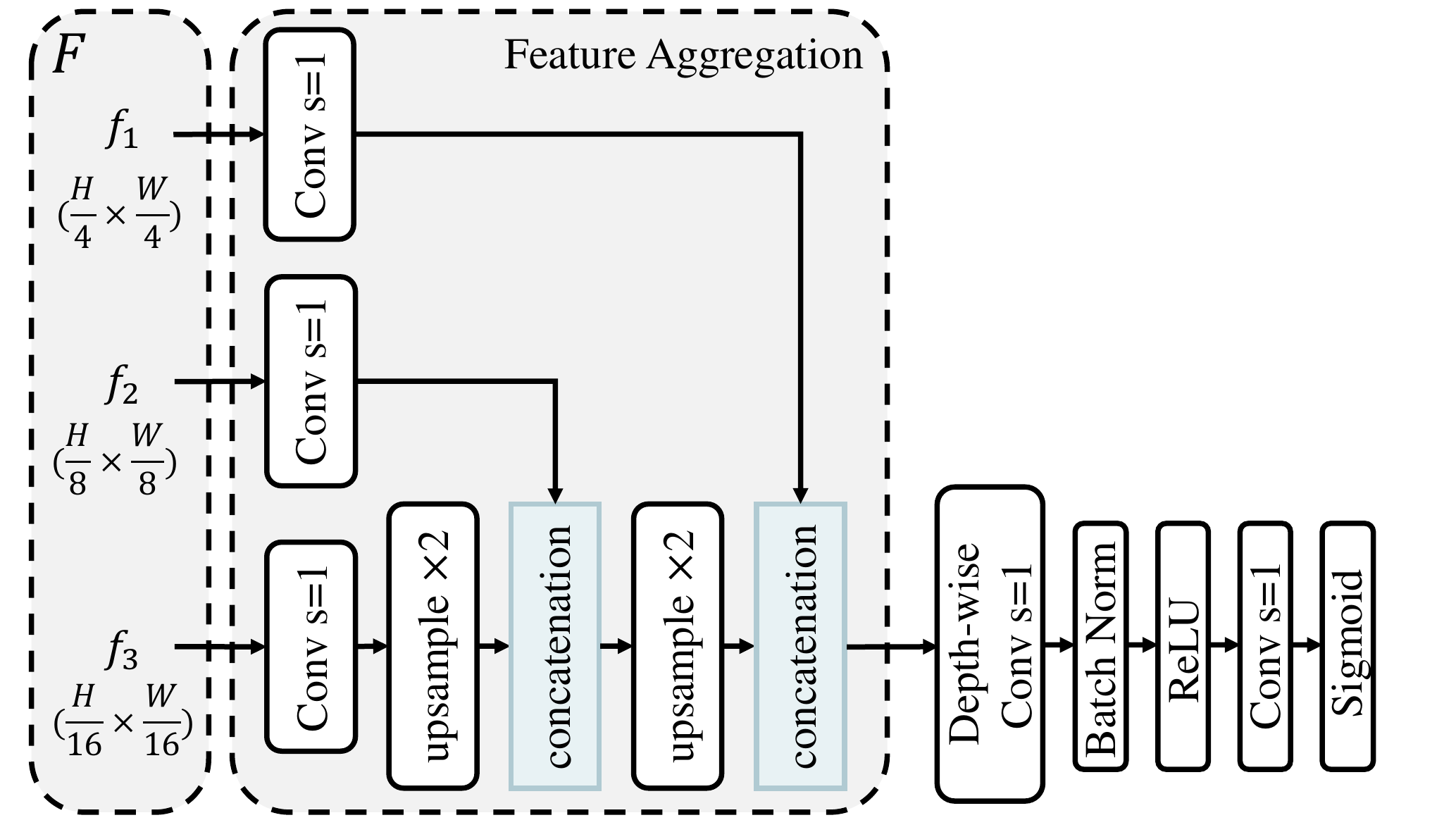}}

    \caption{
    The architecture of information selector with feature aggregation. 
    }
\label{fig:information_selector_feature_aggregation}
\end{figure}

The goal of the information selector is to generate compression prompts that adaptively assign importance factors to each location condition on task requirements and feature contents. 
Since we choose to use the features extracted at stage 1 of Swin Transformer as general features, its network depth is shallow, so the features contain less semantic information. 
Naturally, we can take use of multi-scale features that contain both detailed spatial layout information and high-level semantic information as input of the information selector to generate more proper compression prompts. 
Figure \ref{fig:information_selector_feature_aggregation} illustrates the architecture of the information selector with feature aggregation.
Experimental results demonstrate the effectiveness of our proposed simple and lightweight information selector with feature aggregation.

\section{Comparison with SOTA ICM Methods}
\begin{figure}[t]
  \centerline{\includegraphics[width=1.0\linewidth]{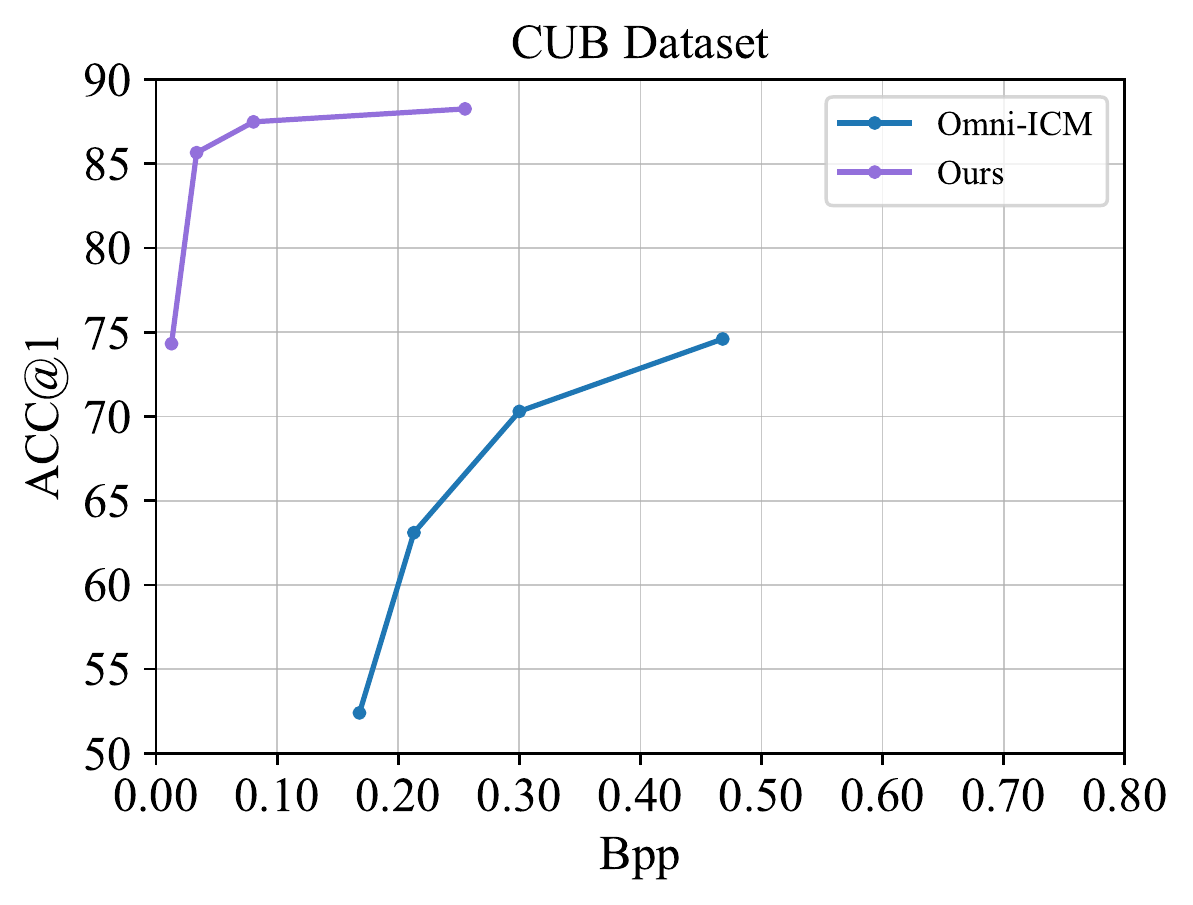}}
    \caption{
    Comparison with Omni-ICM\cite{feng2022image} on CUB-200-2011\cite{wah2011caltech}.
    }
\label{fig:cls_com}
\end{figure}

\begin{figure}[t]
  \centering
\includegraphics[scale=0.3]{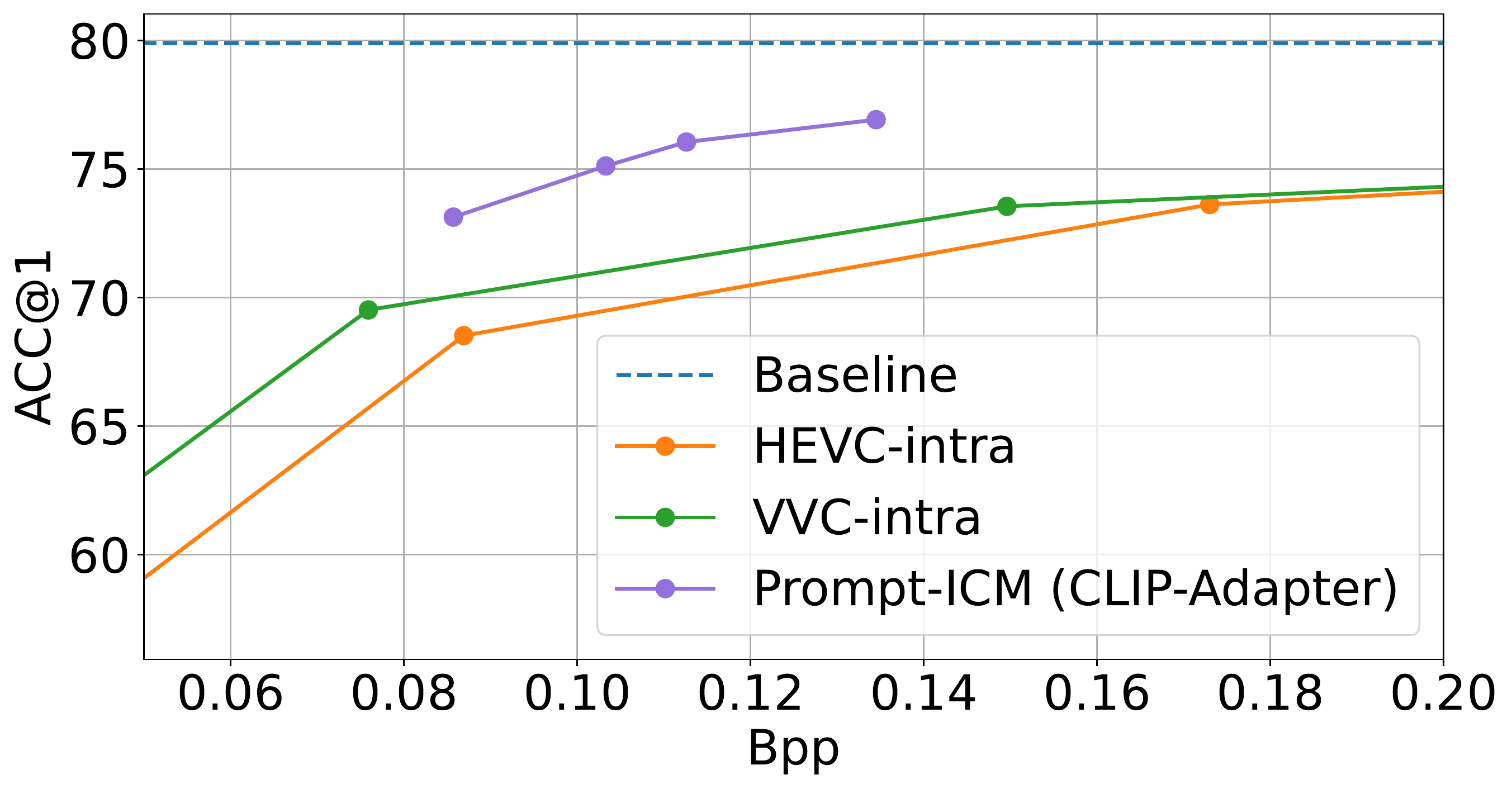}
\caption{
% Classification results on Stanford-Car Dataset by using Prompt-ICM (CLIP-Adapter).
Results based on CLIP-Adapter on Stanford-Car.
}
  \label{fig:tuing}
\end{figure}

\begin{figure}[htbp]
  \centering
\includegraphics[scale=0.85]{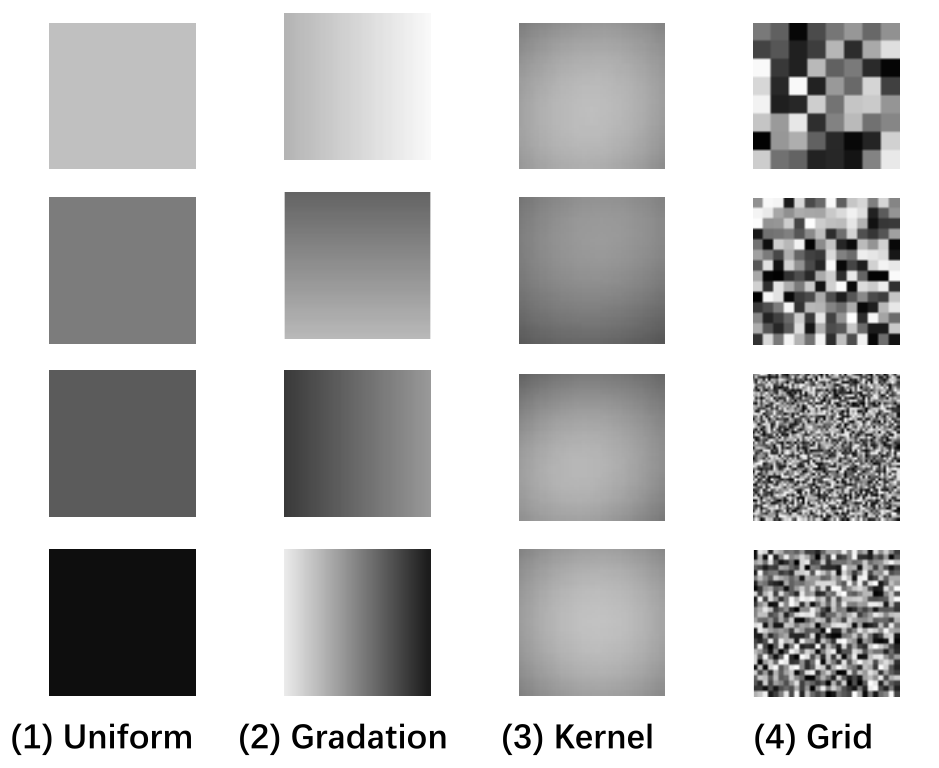}
\caption{
% Classification results on Stanford-Car Dataset by using Prompt-ICM (CLIP-Adapter).
Examples of manual compression prompts.
}
  \label{fig:qmap}
\end{figure}

\begin{figure*}[htbp]
  \centerline{\includegraphics[width=1.0\linewidth]{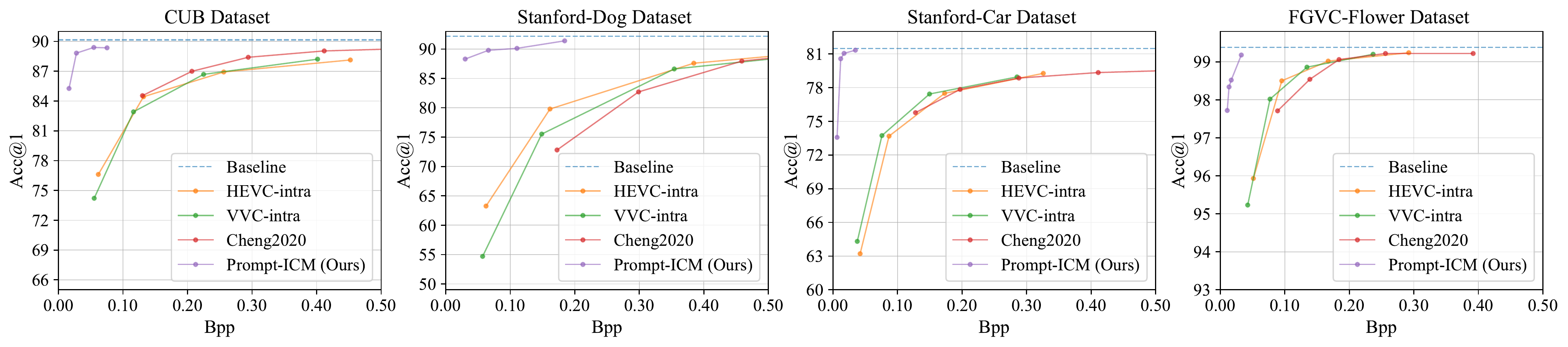}}
    \caption{
    Classification results on different datasets under various bitrates by using ViT-B as the backbone.
    }
\label{fig:Vit_res}
\end{figure*}
\begin{figure*}[htbp]
  \centerline{\includegraphics[width=1.0\linewidth]{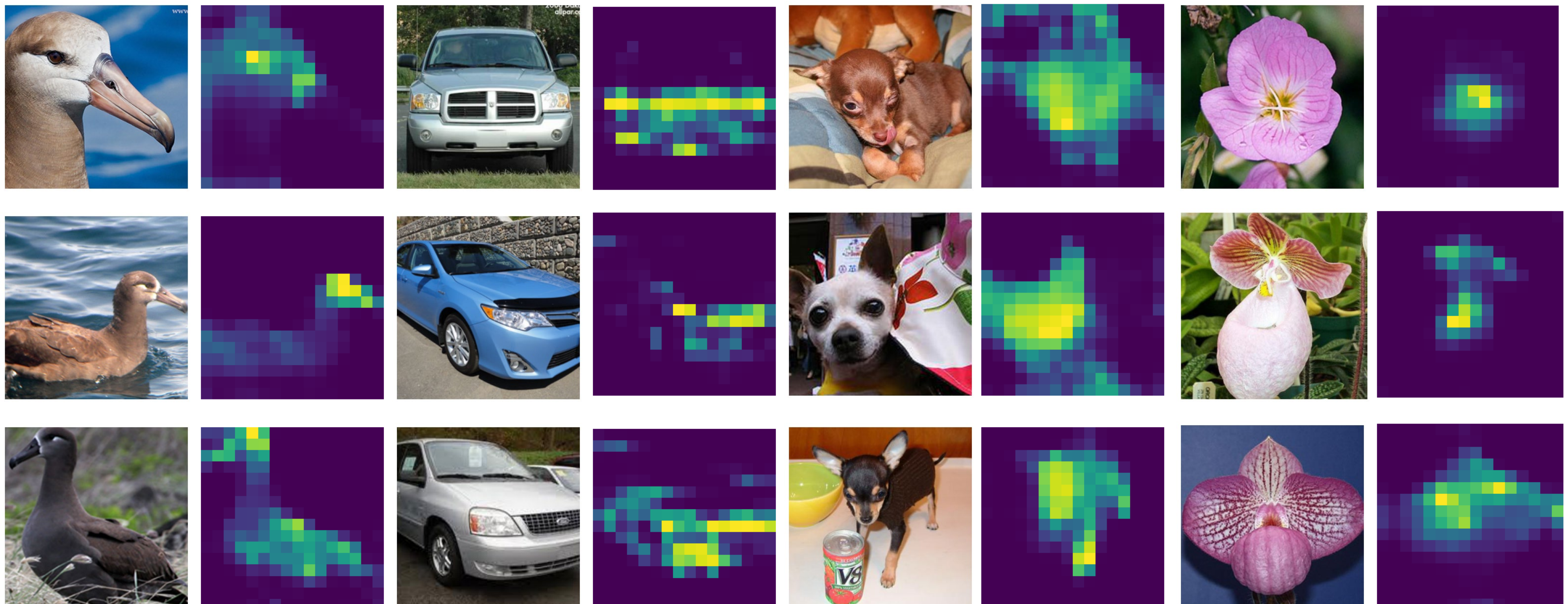}}
    \caption{
    Visualization of compression prompts by ViT-based Prompt-ICM on different datasets. From left to right, the corresponding datasets are CUB-200-2011, Stanford Cars, Stanford Dogs, and Oxford Flowers. Best viewed in color.
    }
\label{fig:supp_vis}
\end{figure*}

In addition to the comparison with dense prediction tasks, we compare our proposed method with Omni-ICM \cite{feng2022image} on CUB-200-2011\cite{wah2011caltech} fine-grained classification task. 
Our Prompt-ICM can achieve far superior performance than Omni-ICM, which shows that our framework has robustness on both image classification and dense prediction tasks. 
We infer that the main reason for the failure of Omni-ICM is because the features learned by contrastive learning in \cite{feng2022image} cannot transfer well to fine-grained classification tasks. 
However, our framework uses more general features, and the proposed task-driven prompts can help us better transfer to downstream tasks, thus obtaining a satisfying performance.

\section{Extension of Task-Adaptive Prompts}

Note that the the task-adaptive prompts in Prompt-ICM is not constrained to any specific prompt tuning techniques. In the main text, we choose VPT\cite{jia2022visual} and Pro-Tuning\cite{nie2022pro} as our instantiation choices. 
Additionally, we have also instantiated the task-adaptive prompts with CLIP-Adapter \cite{gao2021clip} to exhibit the versatility of our framework.
As illustrated in Figure \ref{fig:tuing}, our Prompt-ICM framework retains its superiority over other methods, which demonstrate the adaptability and compatibility of our framework.

\section{Manual Compression Prompts}
The manual prompts mentioned in Section 4.2 of the main text are visualized in Figure \ref{fig:qmap}. Specifically, during the training stage of the codec, to ensure the variety of possible compression prompts, we randomly generate each instance in a mini-batch by using one of the four different ways (1) a uniform map (2) a gradation map between two randomly selected values (3) a kernel density estimation map of a Gaussian mixture with random mean, variance, and a number of mixtures (4) a map consisting of various blocks in a grid manner. 
During the inference stage, to achieve each point in the rate-distortion curve discussed in Section 4.4.1, we employ uniform maps with a range of [0, 1] as compression prompts, which were set manually.

\section{Extension to ViT}
\myparagraph{Feature Extraction and Feature Compression}. We also extend our Prompt-ICM to Vision Transformers (ViT) \cite{dosovitskiy2020image}. 
More specifically, with a normal ViT consisting of 12 self-attention blocks, we take features extracted at block 6 as the general features. 
At the downstream transferring stage, the extracted features are fed into the information selector to generate the compression prompts.
Then, features and compression prompts are input into the feature compression model. 
% The architecture of compression model is nearly the same as swin's, with only all the up-sample and down-sample convolution
The architecture of the compression model for ViT is almost the same as that of Swin. The only difference is that the stride of all convolutions whose original stride is not 1 is changed to 1, since the ViT features are already 16x down-sampling. 
As for task-adaptive prompts, we follow the visual prompt tuning (VPT)\cite{jia2022visual} and take experiments to evaluate the effectiveness on image classification. We conduct experiments on four image classification datasets including CUB-200-2011\cite{wah2011caltech}, Stanford Cars\cite{gebru2017fine}, Stanford Dogs\cite{khosla2011novel}, and Oxford Flowers\cite{nilsback2008automated}.

\myparagraph{Experimental Results}. 
As shown in Figure \ref{fig:Vit_res}, Prompt-ICM can extend well to normal Vision Transformer architecture and substantially outperforms the compared methods. It can be inferred that Prompt-ICM are not limited to a certain backbone, and can achieve excellent performance.

\myparagraph{Compression Prompts Visualization}.
For the ViT-based model, Figure \ref{fig:supp_vis} shows the compression prompts of the four datasets. As we can see, essential patterns for recognition are allocated by more importance. 
For example, the heads of birds, cars and dogs are more important to distinguish the image compared to other patterns, while the flower bud is regarded as the key information for flower classification.

% \section{Further Ablation Study of Task-Adaptive Prompts}

% \fry{Add ablation about prompt tuning here.}

% \fry{
% % During the downstream transferring stage, only parameters of information selector and task-adaptive prompts are tuned, which is mentioned in L503-507 of the main text. 
% The required details of prompts are in Sec-3.4 and Sec-3.5. As mentioned in Line503-507, the parameters of information selectors and task-adaptive prompts need to be tuned in the transferring stage. 
% % And we supplement further ablations in Figure 11, which shows that our method are not significantly sensitive to hyper-parameters of prompt tuning.
% More ablations are conducted and shown in Fig \ref{fig:cls_swin_b_rate_acc}, where the consistent performance indicates Prompt-ICM is not sensitive to hyper-parameters.
% }

% \begin{figure}[htbp]
%   \centering
%   \begin{subfigure}[b]{0.23\textwidth}
%     \includegraphics[width=\textwidth]{figs/rebuttal/depth.pdf}
%     \caption{}
%   \end{subfigure}
%   \begin{subfigure}[b]{0.23\textwidth}
%     \includegraphics[width=\textwidth]{figs/rebuttal/length.pdf}
%     \caption{}
%   \end{subfigure}
%   \caption{The ablation studies of tuning on the CUB dataset. 
%   left: various stages using prompt; right: various length of prompt.}
%   \label{fig:cls_swin_b_rate_acc}
% \end{figure}

\section{Limitation}
The current state of development in parameter efficient tuning (PET) techniques constrains the upper boundary of Prompt-ICM's performance for dense prediction tasks. Nevertheless, as the PET field advances, this limitation is anticipated to be mitigated.

\end{document}